\documentclass[11pt]{article}

\usepackage[preprint]{acl}

\usepackage{times}
\usepackage{latexsym}
\usepackage{enumitem}
\usepackage{setspace}
\usepackage[T1]{fontenc}
\usepackage[utf8]{inputenc}

\usepackage{microtype}

\usepackage{inconsolata}
\usepackage{amsmath}
\usepackage{multicol}
\usepackage{multirow}
\usepackage{booktabs}
\usepackage{subcaption}
\usepackage[most]{tcolorbox}
\usepackage{enumitem}

\usepackage{graphicx}

\title{Video2Code: Generating Interactive Webpages from UI Videos via Action-Aware Revisit}

\author{
\textbf{Mingde Xu}$^{1,3}$\thanks{\ \ Core contributors.}, 
\textbf{Zhen Yang}$^{2,3}$\footnotemark[1]\thanks{\ \ Corresponding author.}, 
\textbf{Yan Wang}$^3$, 
\textbf{Yu Wang}$^3$, 
\textbf{Xijun Liu}$^{3,4}$, 
\textbf{Zijun Dou}$^{2,3}$, 
\\
\textbf{Wenyi Hong}$^{2,3}$, 
\textbf{Xiaotao Gu}$^3$, 
\textbf{Bin Xu}$^2$, 
\textbf{Jie Tang}$^2$\footnotemark[2] \\
$^1$University of Waterloo,
$^2$Tsinghua University,
$^3$Zhipu AI,
$^4$Beihang University
}

\newcommand{\vpara}[1]{\vspace{0.5em}\noindent\textbf{#1}\quad}

\begin{document}
\maketitle
\begin{abstract}

UI videos provide a natural input for generating interactive webpages, as they capture both webpage appearance and action-triggered state transitions. However, directly applying video-capable vision-language models to this task remains insufficient. Existing models typically rely on sparse sampling or compressed temporal representations, which may miss short action boundaries and break the state-action-state transitions needed to implement webpage behavior. We formulate UI video-to-code generation as executable state-transition recovery from interaction videos, and identify this failure mode as state-transition misalignment.
We introduce \textbf{Video2Code}, an action-aware video-to-code approach for recovering executable UI state transitions. Rather than allocating the visual budget uniformly across the video, Video2Code first performs coarse video understanding to locate action-critical regions, then invokes a temporal clipping tool to revisit these regions at higher temporal resolution before generating HTML/CSS/JavaScript code. We instantiate Video2Code with action-aligned video-code supervision and evaluate it under both visual and functional criteria. Experiments show that Video2Code substantially strengthens the underlying open-source model for UI video-to-code generation, improving functional correctness over direct video observation, especially on dense multi-step interactions.

\end{abstract}

\section{Introduction}

% VLMs 发展非常迅速了，已经支持了video的输入，但是video输入更多是连续信息密度的输入。video包含很多片段信息的时候，他们的抽帧逻辑会miss掉里面很多的关键信息。比如说video里面包含交互信息，我们想要理解交互功能以及希望复刻这个网站的时候。

% As shown in Figure 1, 现有的VLMs 尤其是闭源 & 开源 video understanding & code generation 

% 论证一下我们做这个事情的必要性 & 确实有实验结果支撑

% Framework 比较好？还是method？

% As shown in Figure 2, 画我们自己的relook/流程

% In this work, *****我们自己的工作

% \yz{Core idea: Video2Code is an action-aware video-to-code approach for executable state-transition recovery from UI interaction videos. Its core mechanism is Action-Aware Revisit; data/training is only the instantiation that makes this approach trainable and evaluable.}

% \yz{Problem: UI video-to-code generation
% Key challenge: state-transition misalignment
% Core method: Action-Aware Revisit
% Support: action-aligned video-code supervision
% Evaluation: functional correctness / interaction coverage}

\begin{figure*}[t!]
    \centering
    \includegraphics[width=0.96\linewidth]{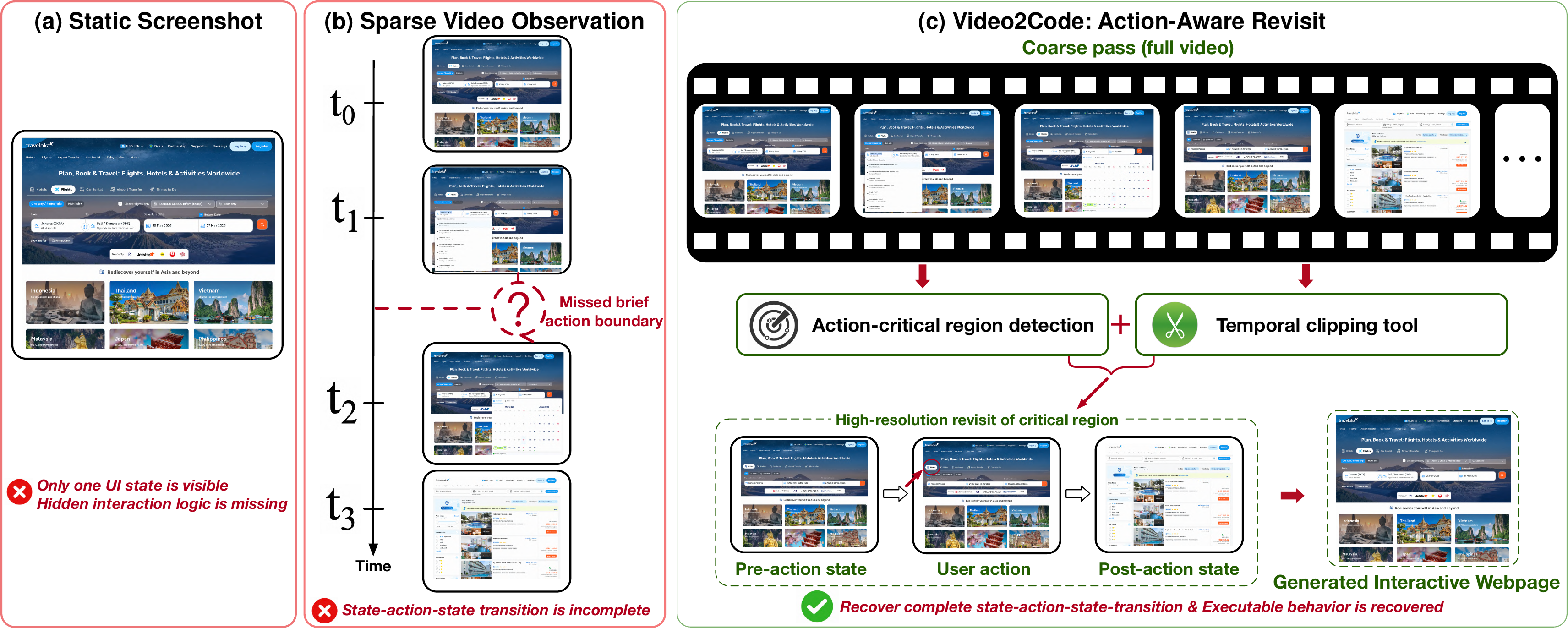}
    % \vspace{-2mm}
    \caption{Motivating example of UI video-to-code generation. Static screenshots capture isolated UI states, while sparse video observation may miss short action boundaries. Video2Code revisits action-critical regions to recover complete state-action-state transitions for executable webpage behavior.}
    \label{fig:motivating_example}
    \vspace{-4mm}
\end{figure*}

Vision-language models (VLMs) have recently shown strong capabilities in multimodal understanding and code generation~\cite{liu2023visual,wang2024qwen2,wang2024cogvlm,bai2025qwen2,hong2025glm,zhu2025internvl3}. This progress has led to growing interest in UI-to-code generation, where models produce executable front-end code from webpage screenshots or design images~\cite{si2024design2Code,yun2024web2code,jiang2025screencoder,gui2025uicopilot,gui2025webcode2m,xu2025webvia}. However, most existing methods still operate on static screenshots, with the goal of reconstructing the layout and visual appearance of a webpage. Such a formulation captures what an interface looks like, but provides little information about how it behaves under user interaction.

% Modern webpages often reveal their behavior only through interaction. A screenshot may show a collapsed menu or the initial viewport of a long page, but it does not capture what happens after a click, selection, or scroll. Even multiple screenshots remain ambiguous when the action order and transition process are missing. Thus, interactive webpage generation requires recovering not only individual UI states, but also the action-conditioned transitions between them. UI videos provide a natural input for this setting because they record the temporal evidence needed to infer such transitions. As shown in Figure~\ref{fig:motivating_example}, the gap between static and interactive UI-to-code generation lies not only in reconstructing visual appearance, but also in recovering the behavior triggered by user actions. A generated page may match the initial interface while failing to reproduce the transitions after clicks, selections, or scrolling.

Modern webpages often reveal their behavior only through interaction. A screenshot may show a collapsed menu or an initial viewport, but it cannot specify what happens after a click, selection, or scroll. Even multiple screenshots remain ambiguous when the action order and transition process are missing. UI videos therefore provide a natural input for interactive webpage generation, as they record both interface states and the temporal evidence connecting them. As shown in Figure~\ref{fig:motivating_example}, the key gap is not only visual reconstruction, but also recovering the behavior triggered by user actions.

However, video input alone is not sufficient. Current video-capable VLMs usually process videos with sparse frame sampling~\cite{cheng2024videollama,chen2024sharegpt4video,cheng2025scaling,liu2025bolt,zhang2026one} or compressed temporal representations~\cite{li2024llava,zhang2025videollama,zhang2024llava,li2024llama}. While effective for many natural videos, such budget-limited observation is fragile for UI interaction videos, where code-relevant evidence is concentrated around action boundaries. For executable webpage generation, the key evidence is not an isolated informative frame, but the complete state-action-state transition around a user operation.

We therefore formulate UI video-to-code generation as \emph{executable state-transition recovery} from interaction videos. Under budget-limited video observation, a model may suffer from \emph{state-transition misalignment}: the sampled observations fail to preserve the complete state-action-state transition needed to implement webpage behavior. To address this challenge, we introduce \textbf{Video2Code}, an action-aware video-to-code approach for recovering executable UI state transitions. Rather than allocating observation uniformly across the video, Video2Code first performs a coarse pass to locate action-critical regions, then invokes a temporal clipping tool to revisit these regions at higher temporal resolution before generating executable HTML/CSS/JavaScript code. This coarse-to-fine design preserves global context while recovering the local transition evidence needed for behavior-faithful webpage generation.

% We instantiate this approach with action-aligned video-code supervision, where each UI interaction video is paired with temporal action anchors and an executable webpage implementation. This supervision teaches the model both when to revisit action-critical regions and how to translate recovered state transitions into front-end code. We evaluate Video2Code under both visual and functional criteria. Beyond measuring visual similarity, we use browser-based interaction verification to test whether the generated webpage reproduces the behavior demonstrated in the input video. Experiments show that Video2Code improves functional correctness and interaction coverage over single-pass video observation, especially on dense multi-step interactions.

We instantiate Video2Code with action-aligned video-code supervision. Each UI interaction video is paired with temporal action anchors and an executable webpage implementation, enabling the model to learn both revisit planning and code generation from recovered transition evidence. We evaluate Video2Code under both visual and functional criteria, including browser-based interaction verification. Experiments show that Video2Code improves functional correctness and interaction coverage over single-pass video observation, especially on dense multi-step interactions.

Our contributions are summarized as follows:

\begin{itemize}
    \item We formulate UI video-to-code generation as executable state-transition recovery from interaction videos, and identify \emph{state-transition misalignment} as a key challenge of budget-limited video observation.
    \item We propose \textbf{Video2Code}, an action-aware video-to-code approach that uses a temporal clipping tool to revisit action-critical regions at higher temporal resolution before generating executable webpage code.
    \item We instantiate Video2Code with action-aligned video-code supervision and show that it improves functional correctness and interaction coverage over single-pass video observation.
\end{itemize}

\section{Video2Code}
\label{headings}

% Our goal is to enable VLMs to identify and extract relevant information from long videos with uneven information density, where critical content may be obscured by temporal gaps between frames. 
% To serve this purpose, we introduce the Clip-and-Revisit method, where the model invokes external tools when encountering aforementioned situation, and refines it's understanding based on the returned results.

% 2. Method: 解决方案：为了训推一致，用Function call对于网页video进行切片，二次浏览，保证关键信息。我们对于切片的具体定义，什么样的信息需要切（点击，输入，选择，滚动）（主Method） 

\begin{figure}[t!]
    \centering
    \includegraphics[width=\linewidth]{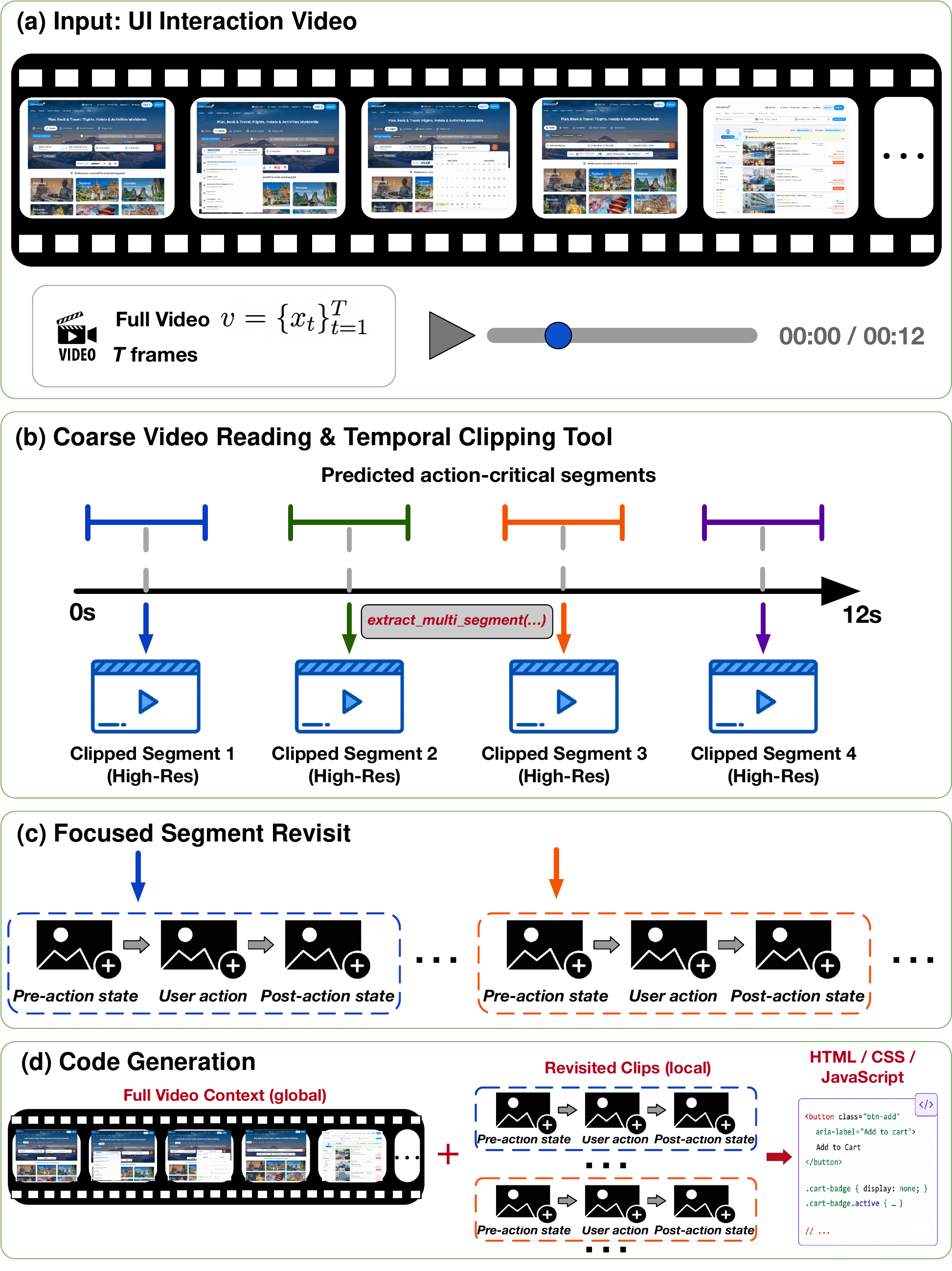}
    % \vspace{-2mm}
    \caption{Overview of Video2Code. Action-Aware Revisit locates action-critical segments, revisits them with a temporal clipping tool, and generates executable webpage code from global and local evidence.}
    \label{fig:video2code_inference}
    \vspace{-4mm}
\end{figure}

In this section, we present \textbf{Video2Code}, an action-aware video-to-code approach for generating executable front-end code from UI interaction videos. As shown in Figure~\ref{fig:video2code_inference}, Video2Code centers on \emph{Action-Aware Revisit}: the model first performs a coarse pass over the full video to locate action-critical regions, then invokes a temporal clipping tool to revisit these regions at higher temporal resolution before generating the final webpage implementation.

\subsection{Problem Setup}

Given a UI interaction video $v=\{x_t\}_{t=1}^{T}$, where $x_t$ denotes the frame at time $t$, Video2Code aims to generate an executable webpage implementation $y$, including HTML, CSS, and JavaScript. We view $v$ as demonstrating a latent sequence of action-conditioned UI transitions:
\[
\mathcal{Z}(v)=\{(s_i, a_i, s_{i+1}, \tau_i)\}_{i=1}^{N},
\]
where $s_i$ and $s_{i+1}$ are the UI states before and after a user action $a_i$, and $\tau_i=(t^i_s,t^i_e)$ denotes the temporal window containing the transition.

A behavior-faithful implementation should reproduce both the visual states and the executable behavior: executing the same action $a_i$ on the generated webpage should lead to a state matching $s_{i+1}$. Thus, the key observation unit for UI video-to-code generation is not an isolated frame, but a complete state-action-state transition around a user operation.

\subsection{Temporal Clipping Tool}

To support action-aware revisit, Video2Code uses a temporal clipping tool that extracts model-specified video segments for focused observation. Given an input UI interaction video $v$, the model can issue a tool call
\[
a_t=\texttt{extract\_multi\_segment}(v,\hat{\tau}_t),
\]
where $\hat{\tau}_t=\{(\hat{t}^{j}_{s},\hat{t}^{j}_{e})\}_{j=1}^{K}$ denotes a set of temporal segments predicted by the model at step $t$. Each segment $(\hat{t}^{j}_{s},\hat{t}^{j}_{e})$ specifies a video interval that is expected to contain action-critical transition evidence. Executing the tool call returns the corresponding clipped video observations:
\[
o_t = E(a_t)=\{c_1,c_2,\ldots,c_K\},
\]
where each $c_j$ is the clipped segment associated with $(\hat{t}^{j}_{s},\hat{t}^{j}_{e})$.

The returned clips are incorporated into the model context for subsequent reasoning:
\[
\mathcal{H}_{t+1}=\mathcal{H}_t \oplus o_t,
\]
where $\mathcal{H}_t$ denotes the current multimodal context and $\oplus$ denotes context augmentation. Since each clipped segment covers a much shorter temporal range than the original video, the model can inspect it at a higher temporal resolution without uniformly densifying the full video. The tool itself does not decide which regions are important; it only executes the temporal intervals specified by the model.

% This enables Video2Code to recover transient evidence around user actions, including the pre-action state, the action execution process, and the resulting post-action state.

% The clipping tool itself does not decide which regions are important; it only executes the temporal intervals specified by the model. 
% Thus, region selection remains part of the model's reasoning process, while the tool provides controllable access to higher-resolution temporal evidence for action-aware revisit.

\subsection{Action-Aware Revisit}

% Video2Code performs video-to-code generation through a two-round revisit process. The first round builds a global understanding of the interaction video and predicts where executable state transitions are likely to occur. The second round uses the temporal clipping tool to revisit these predicted regions at higher temporal resolution, allowing the model to recover local transition evidence before generating code.

Video2Code follows a two-round revisit process: it first builds a coarse understanding of the full interaction video to locate action-critical regions, and then revisits these regions with the temporal clipping tool to recover local transition evidence before code generation.

\vpara{Coarse Video Reading.} In the first round, the model observes the full UI interaction video at a coarse temporal resolution to infer the global interaction flow and produce a revisit plan. Specifically, it predicts action-critical temporal segments $\hat{\tau}$ that are likely to contain executable transition evidence around user operations such as clicks, text inputs, selections, and scrolls. The predicted segments are then passed to the temporal clipping tool for focused extraction.

% In the first round, the model observes the full UI interaction video at a coarse temporal resolution. Rather than directly generating code from this incomplete observation, the model first infers the global interaction flow and produces a revisit plan. Specifically, it predicts a set of action-critical temporal segments $\hat{\tau}_t$ that are likely to contain complete state-action-state transitions. These segments typically correspond to user operations such as clicks, text inputs, selections, and scrolls, where a short temporal window may determine the executable behavior of the webpage. The predicted segments are then passed to the temporal clipping tool for focused extraction.

\vpara{Focused Segment Revisit.} In the second round, the returned clips are reintroduced into the model context as focused observations. The model combines the global context from the coarse pass with the local evidence from the revisited clips to generate the final HTML/CSS/JavaScript implementation. This process preserves global interaction structure while recovering local transition evidence needed for behavior-faithful code generation.

% In the second round, the extracted clips are reintroduced into the model context as focused observations. Because these clips cover only selected temporal windows, the model can inspect the local state changes around each action at a higher temporal resolution. The model then combines the global context from the coarse pass with the detailed evidence from the revisited clips to generate the final HTML/CSS/JavaScript implementation. This design avoids uniformly increasing the frame budget over the entire video, while preserving the state-action-state evidence required for behavior-faithful code generation.

\subsection{Defining Action-Critical Segments}

A key question in Action-Aware Revisit is what should be revisited. For UI video-to-code generation, an action-critical segment is not simply a visually salient frame or an informative video moment. Instead, it should provide sufficient temporal evidence for recovering an executable transition. Thus, we define an action-critical segment as a video interval that covers three elements: the UI state before a user operation, the operation itself, and the resulting UI state after the operation. This definition ensures that the revisited segment provides the local state-action-state evidence needed for generating event handlers, state updates, and dynamic UI behavior.

\vpara{Pure Action Segments.} A pure action segment contains a single user operation whose effect can be understood locally, such as a click, text input, selection, or scroll. For such actions, the segment should include the pre-action state, the action execution process, and the post-action state. For example, when a click opens a dropdown menu, the segment should cover the collapsed menu, the click operation, and the expanded menu state. This allows the model to infer both the triggering event and the UI change that should be implemented in code.

\vpara{Action-Scroll Segments.} Some webpage behaviors cannot be recovered from isolated actions because the resulting state is revealed only after a subsequent scroll. For example, clicking a navigation link may load new content, while scrolling exposes the newly loaded section. If the click and scroll are clipped independently, the model may miss the contextual dependency between the triggering action and the visible post-action state. Therefore, we group such action-scroll sequences into a single segment when the scroll is necessary to observe the result of the preceding action.

This transition-centered definition distinguishes Video2Code from general frame or clip selection. The goal is not merely to retain visually informative content, but to preserve complete transition evidence that can be translated into executable webpage behavior.

\section{Action-Aligned Supervision and Training}

\begin{figure}[t!]
    \centering
    \includegraphics[width=\linewidth]{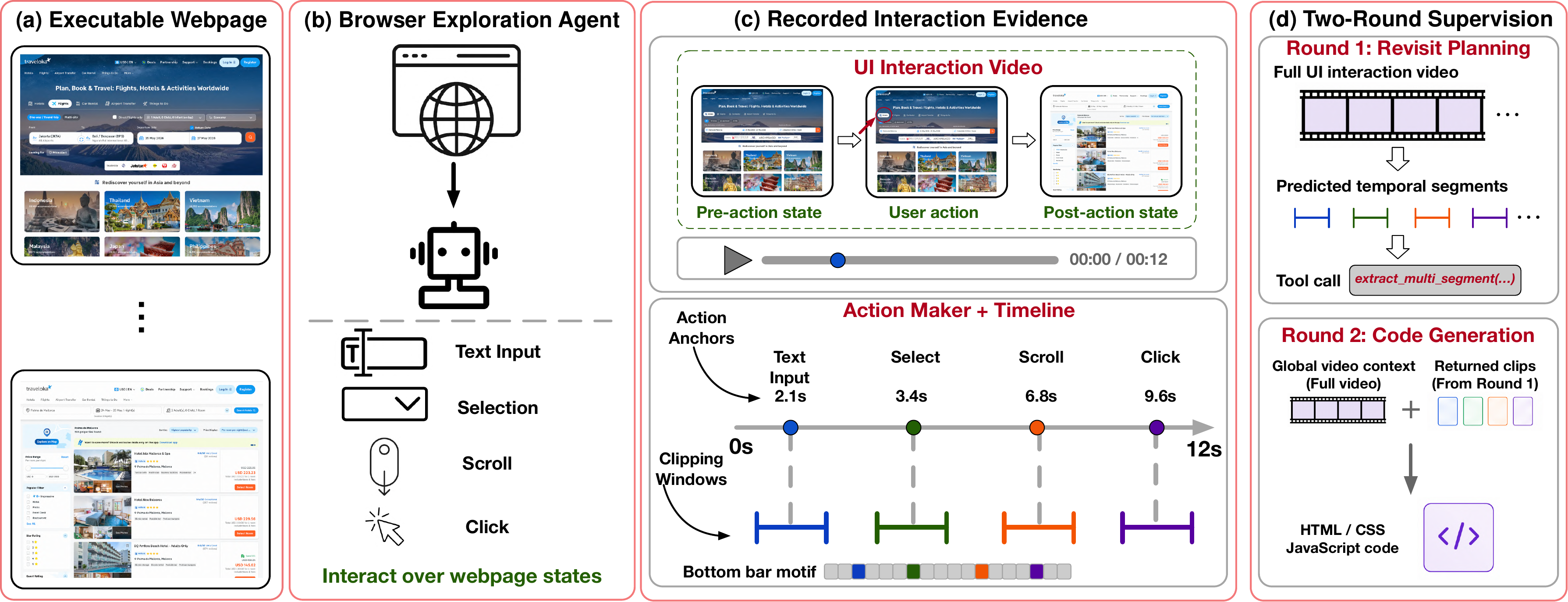}
    % \vspace{-2mm}
    \caption{Action-aligned supervision for Video2Code. Interaction videos and action timelines supervise temporal clipping, while webpage implementations supervise executable code generation from the global context and returned clips.}
    \label{fig:supervision_pipeline}
    \vspace{-4mm}
\end{figure}

To train Video2Code, we construct action-aligned supervision that follows the same two-round structure as Action-Aware Revisit.
Figure~\ref{fig:supervision_pipeline} illustrates the overall supervision and training pipeline. 
% Unlike a standard video-code pair that only provides a video-level input and a target implementation, our supervision explicitly aligns webpage interaction trajectories with temporal clipping and code generation. The first round teaches the model to identify action-critical regions and invoke the clipping tool, while the second round teaches it to generate executable code from the global video context and the returned clips.

\subsection{UI Interaction Video Collection}

We collect UI interaction videos from executable webpages through automated browser exploration. Given a webpage URL or an HTML file, a web exploration agent interacts with the page by executing common user operations such as clicks, text inputs, selections, and scrolls. The exploration is performed over webpage states to expose visual states and action-triggered transitions, while the browser viewport is recorded as a UI interaction video.

To obtain temporal anchors for user actions, we inject a lightweight visual marker into the webpage during recording. Specifically, a 12-pixel-high bar is added to the bottom of the page and changes color when an action is executed. This marker does not affect the webpage layout or interaction logic, but provides an explicit signal for localizing action boundaries. We then derive action timelines that specify when each operation occurs and which temporal windows should be considered for clipping.

For each webpage, the collected data consists of a UI interaction video, an action timeline with temporal anchors, and the corresponding webpage implementation. 
The video provides the input observation, the timeline provides supervision for action-aware temporal clipping, and the implementation provides the target executable code.

\subsection{Constructing Action-Aligned Video-Code Supervision}

Given a UI interaction video, its action-aligned timeline, and the corresponding webpage implementation, we construct supervision that follows the same two-round structure as Video2Code inference. 

% The goal is to align training with the revisit-based generation process: the model first learns to plan temporal clipping from the full video, and then learns to generate executable code conditioned on the global context and the returned clips.

\vpara{Round 1: Revisit Planning.} The first round teaches the model to locate action-critical regions from the full interaction video. We convert the action-aligned timeline into target tool-call supervision. Specifically, each action anchor is expanded into a temporal clipping window that covers the pre-action state, the action execution process, and the post-action state. The model is trained to predict these temporal windows and invoke the temporal clipping tool accordingly. This supervision encourages the model to focus on executable transition evidence rather than treating all video moments as equally informative.

\vpara{Round 2: Code Generation after Revisit.} The second round teaches the model to generate executable webpage code after the clipped segments are returned. For each example, the model receives the global video context together with the revisited clips extracted in the first round. The target output is the corresponding webpage implementation, including HTML, CSS, and JavaScript. This supervision encourages the model to ground code generation in both the overall interaction flow and the local transition evidence recovered through revisit.

% \vpara{Quality Filtering.} We further filter the constructed examples to improve supervision quality. The filtering checks whether the action timeline correctly covers interaction boundaries, whether the returned clips contain sufficient transition evidence, and whether the target implementation is executable and consistent with the observed webpage behavior. After filtering, we obtain the final \textit{WebVidCoding} dataset, consisting of action-aligned video-code examples for training Video2Code.

\vpara{Annotation and Filtering.} We use Gemini-3.1-Pro to annotate the two-round supervision. 
For the first round, the action-aligned clipping timelines are converted into target tool-call trajectories, including the expected temporal segments and tool calls. For the second round, each clipped segment is represented as a sequence of frames and used to produce the corresponding reference webpage implementation. After annotation, we apply a function judge filtering step to remove low-quality examples. The filtering mainly checks code quality and reasoning/tool-call consistency. After filtering, we obtain \textit{WebVidCoding}, a dataset of 12K fully annotated action-aligned video-code pairs for training Video2Code. Further details of \textit{WebVidCoding} construction are provided in Appendix~\ref{app:data_construction}.

\subsection{Training}

We fine-tune Video2Code on \textit{WebVidCoding} with supervised fine-tuning, using GLM-4.6V-100B as the base model. Each training example follows the two-round format of Action-Aware Revisit: the model first predicts tool calls for action-critical temporal segments from the full video, and then generates the executable webpage implementation conditioned on the returned clips. This training format matches the inference-time revisit process. Training details are provided in Appendix~\ref{app:training_details}.

\section{Experiments}

% setup: 真实网页88个，合成网页50个 真实网页：自己找了100个现实网页，手动录制网页交互视频，进行annotation，表基础交互发生的时间点，筛选掉一部分，筛选标准是啥：交互过多过少，保留了88个。合成网页：用我们的构造训练数据的 webvia-video-agent进行探索，过滤，视频重构（把原本探索中冗余的剪切掉），得到100个合成网页上的交互视频

% evaluation metrics：1， visual similarity 2，functional correctness

% 都是pagewise，每个page内部取平均，然后整体取平均 （因为每个webpage有多个tasks）

% Verify Agent：

% 1. 主实验： 合成和真实网站上的结果

% 2. Ablation:

% 1. tool call -without tool call （有无revisit）

% 2. focused segment revisit中的抽帧策略 （1fps + 3fps）or 1fps ——针对抽下来的视频是怎么抽帧的？长视频（完整视频） + 短视频（切下来的视频） ❌

% 3. 

% 4. 视频长度 

% 5. 合成网页中 scroll 交互速度对于结果的影响，20个样本，快、中、慢

% \yz{todo 5.23 Done!}

In this section, we evaluate Video2Code under both visual and functional criteria. Our experiments examine whether Video2Code can preserve visual states and action-triggered behaviors from UI videos, whether Action-Aware Revisit improves over direct video-to-code generation and strong VLM baselines, and whether the gains hold across real-world, synthetic, and complex interaction scenarios.

% Our experiments are designed to answer three questions: 
% (1) whether Video2Code can generate executable webpages that preserve both visual states and action-triggered behaviors demonstrated in UI videos; 
% (2) whether Action-Aware Revisit improves over direct video-to-code generation and strong video-capable VLM baselines; and 
% (3) whether the improvements remain consistent across real-world webpages, synthetic webpages, and interaction scenarios with different levels of complexity. 

\begin{table*}[t!]
\centering
\small
\caption{Main results on \textit{WebVideo2Code-Bench}. Visual similarity and functional correctness are computed page-wise by first averaging over interaction tasks within each webpage and then averaging over webpages.}
\vspace{-2mm}
\renewcommand{\arraystretch}{1.1}
\begin{tabular}{lcccc}
\toprule
\multirow{2}{*}{Model} 
& \multicolumn{2}{c}{WebVid2Code-Real} 
& \multicolumn{2}{c}{WebVid2Code-Syn} \\
\cmidrule(lr){2-3} \cmidrule(lr){4-5}
& Visual Sim.  & Func. Corr.  
& Visual Sim.  & Func. Corr.  \\
\midrule
\multicolumn{5}{c}{\textbf{Closed-source VLM}} \\
\midrule
Gemini-3.1-Pro & 82.53 & 71.64 & 86.28 & 58.30 \\
Gemini-2.5-Pro & 79.99 & 61.81 & 86.83 & 56.17 \\
Kimi-K2.6 & 86.11 & 73.14 & 92.45 & 71.27 \\
% GPT-5.5 & -- & -- & -- & -- \\
% Claude-Sonnet-4.6  & -- & -- & -- & -- \\
\midrule
\multicolumn{5}{c}{\textbf{Open-source VLM}} \\
\midrule
Qwen3.5-122B-A10B-Thinking & 78.01 & 43.21 & 87.01 & 50.44 \\
Qwen3-VL-235B-A22B-Thinking & 71.75 & 35.29 & 80.80 & 33.06 \\
% InternVL3.5-8B & -- & -- & -- & -- \\
% Mimo-v2.5-Pro & -- & -- & -- & --  \\
\midrule
GLM-4.6V & 70.94 & 29.85 & 79.53 & 53.59 \\ 
Video2Code & \textbf{82.06} & \textbf{61.99} & \textbf{88.53} & \textbf{58.06} \\
\bottomrule
\end{tabular}
\label{tab:main_results}
\vspace{-3mm}
\end{table*}

\subsection{Experimental Setup}

\vpara{Evaluation Models.} We compare Video2Code with closed-source and open-source multimodal models that support native video input and can generate code. Closed-source models include Gemini-3.1, Gemini-2.5, and Kimi-K2.6 accessed via official APIs, while open-source models include representative Qwen3.5~\cite{team2026qwen3} and Qwen3-VL~\cite{bai2025qwen3}. All baselines are prompted to directly generate an executable HTML/CSS/JavaScript webpage from the input UI interaction video. 
We focus on code-capable VLMs rather than video-understanding-only models to avoid conflating UI video understanding with basic code generation ability.

\vpara{Benchmarks.} We construct \textit{WebVideo2Code-Bench} for UI video-to-code generation, consisting of two subsets that cover real-world and controlled webpage scenarios. \textit{WebVideo2Code-Real} contains 85 real-world webpages selected from 100 manually recorded websites. For each website, we operate the webpage in a browser, record an interaction video, annotate user-action timestamps, and filter out recordings with overly sparse, overly dense, or ambiguous interactions. \textit{WebVideo2Code-Syn} contains 76 synthetic webpages built from trajectories collected by an exploration agent. We filter low-quality trajectories and remove redundant or repeated operations to obtain concise benchmark videos. Both subsets cover common operations such as click, text inputs, selections, and scrolls. Each webpage contains multiple interaction tasks, and we report final scores by first averaging over tasks within each webpage and then over webpages. Detailed benchmark construction procedures and statistics are provided in Appendix~\ref{app:benchmark_contruction}.

\vpara{Metrics.} We evaluate generated webpages using two complementary metrics: \textit{visual similarity} and \textit{functional correctness}. Visual similarity measures whether the rendered states of the generated webpage match the corresponding target states in the input video. Functional correctness measures whether the generated webpage can reproduce the expected state changes after executing the demonstrated user actions. Both metrics are computed at the task level and then aggregated page-wise: for each webpage, we first average scores over its interaction tasks, and then average page-level scores over all webpages. This prevents webpages with more tasks from dominating the final results.

\subsection{Browser-Based Functional Verification}

To evaluate whether a generated webpage reproduces the behavior shown in the input video, we perform browser-based interaction verification. Given an annotated interaction video, we first split it into task-wise frame sequences according to the user-action timestamps, where each task corresponds to one operation and its surrounding state transition. For each task, we launch the generated webpage in a browser and provide the verifier with the current webpage screenshot, DOM tree, and the target task frames. The verifier then executes the operation that best matches the demonstrated action, with an action space including click, text entry, selection, and scroll.

During execution, the browser records the resulting webpage trace as a sequence of frames. The verifier compares this execution trace with the target task frames and produces two judgments: whether the expected state change occurs, and how visually similar the resulting state is to the target state. These judgments are used to compute functional correctness and visual similarity, respectively.

For real-world webpages, tasks are executed sequentially following the order of the manually recorded interaction video. To reduce cascading failures, we use an initial-page retry mechanism: when a task cannot be executed from the current state, we reload the generated webpage and retry the same task from the initial page. If the retry succeeds, the new browser state is used for subsequent tasks; otherwise, the task is marked as failed. Further details of the evaluation protocol, task-level scoring, and page-wise aggregation are provided in Appendix~\ref{app:evaluation_metrics}.

\subsection{Main Results}

Table~\ref{tab:main_results} reports the main results on \textit{WebVideo2Code-Bench}, including \textit{WebVideo2Code-Real} and \textit{WebVideo2Code-Syn}. Closed-source VLMs such as Kimi-K2.6 and Gemini-3.1-Pro achieve strong performance, showing that UI video-to-code generation benefits from powerful general-purpose multimodal reasoning. 
However, direct generation with open-source models remains limited, especially in functional correctness. Compared with the GLM-4.6V direct generation baseline, Video2Code improves visual similarity from 70.94 to 82.06 and substantially improves functional correctness from 29.85 to 61.99 on \textit{WebVideo2Code-Real}. On \textit{WebVideo2Code-Syn}, Video2Code also improves over GLM-4.6V from 79.53 to 88.53 in visual similarity and from 53.59 to 58.06 in functional correctness. These results suggest that Action-Aware Revisit and action-aligned supervision help recover executable state transitions across both real-world and controlled webpage scenarios, substantially strengthening the underlying open-source model for behavior-faithful UI video-to-code generation.

\subsection{Ablation Study}

% We conduct an ablation study on \textit{WebVideo2Code-Real} to isolate the effect of Action-Aware Revisit in real-world webpage scenarios. The \textit{w/o Revisit} variant uses the same fine-tuned model as Video2Code, but disables the temporal clipping tool and directly generates code from the coarse video observation. All other settings are kept unchanged. As shown in Table~\ref{tab:ablation}, removing Action-Aware Revisit leads to clear performance drops. Video2Code improves visual similarity from 77.17 to 82.06 and more substantially improves functional correctness from 47.84 to 61.99. The larger gain in functional correctness suggests that focused revisit is especially important for recovering executable transition evidence around user actions. While the coarse video observation can capture the overall webpage layout and interaction flow, it may miss short action boundaries or local state changes that are necessary for implementing event handlers, state updates, and other dynamic behaviors.

We conduct an ablation study on \textit{WebVideo2Code-Real} to isolate the effect of Action-Aware Revisit in real-world webpage scenarios. The \textit{w/o Revisit} variant uses the same fine-tuned model as Video2Code, but disables the temporal clipping tool and directly generates code from the coarse video observation. All other settings are kept unchanged. As shown in Table~\ref{tab:ablation}, removing Action-Aware Revisit leads to clear performance drops. Video2Code improves visual similarity from 77.17 to 82.06, and more substantially improves functional correctness from 47.84 to 61.99. The larger gain in functional correctness suggests that focused revisit is particularly important for recovering executable transition evidence around user actions.

\begin{table}[t!]
\centering
\small
\renewcommand{\arraystretch}{1.1}
\caption{Ablation study of Action-Aware Revisit on \textit{WebVideo2Code-Real}. The \textit{w/o Revisit} variant disables the temporal clipping tool.}
\vspace{-2mm}
\begin{tabular}{lcc}
\toprule
Model 
& \multicolumn{2}{c}{\textit{WebVideo2Code-Real}} \\
\cmidrule(lr){2-3}
& Visual Sim.  
& Func. Corr.    \\
\midrule
Video2Code w/o Revisit & 77.17 & 47.84  \\
Video2Code & \textbf{82.06} & \textbf{61.99} \\
\bottomrule
\end{tabular}
\label{tab:ablation}
\vspace{-3mm}
\end{table}

\subsection{Effect of Frame-Based Observation}

% We further examine whether sparse frame-based observation is sufficient for UI video-to-code generation. For models without native video input, a common alternative is to convert the input video into sampled frames and provide them as multi-image context. 
% We evaluate GPT-5.5 with 1 FPS sampling, which provides relatively dense visual observations without using native video processing or temporal revisit. As shown in Table~\ref{tab:frame_based_baselines}, the frame-based baseline achieves high visual similarity, indicating that sampled frames provide strong cues for reconstructing webpage appearance. However, its functional correctness remains lower than Video2Code, suggesting that visual coverage alone is not sufficient for behavior-faithful webpage generation. Although 1 FPS sampling captures many representative UI states, it can still miss short action boundaries and the precise state-action-state evidence needed to implement executable transitions. These results support the need for action-aware temporal observation in UI video-to-code generation.

We further examine whether sparse frame-based observation is sufficient for UI video-to-code generation. For models without native video input, we convert each input video into sampled frames and provide them as multi-image context. Specifically, we evaluate GPT-5.5 with 1 FPS sampling, which provides relatively dense visual observations without native video processing. As shown in Table~\ref{tab:frame_based_baselines}, this setting achieves high visual similarity, but its functional correctness remains lower than Video2Code. This suggests that sampled frames can provide strong cues for webpage appearance, but may still miss action-critical temporal evidence needed for executable behavior.

\begin{figure*}[t!]
    \centering

    \begin{subfigure}[t]{0.32\linewidth}
        \centering
        \includegraphics[width=\linewidth]{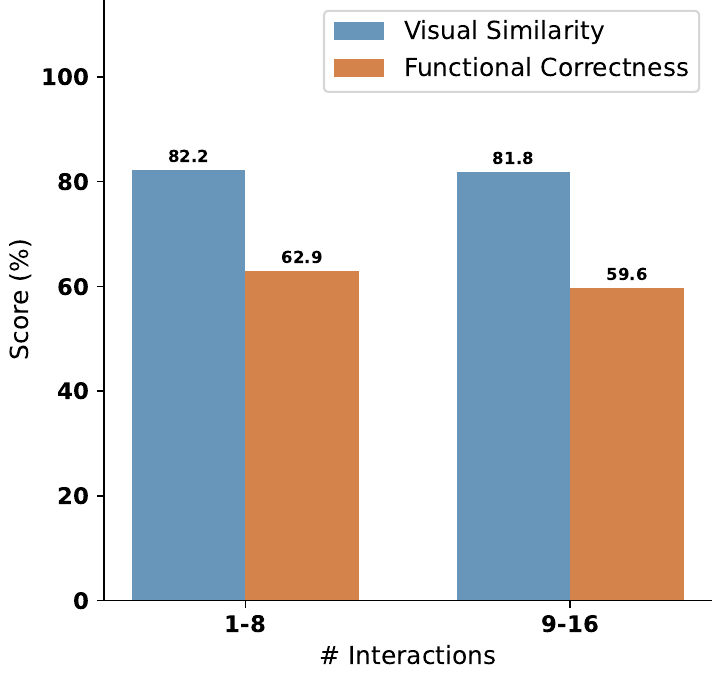}
        \caption{Interaction count}
        \label{fig:analysis_interaction_count}
    \end{subfigure}
    \hfill
    \begin{subfigure}[t]{0.32\linewidth}
        \centering
        \includegraphics[width=\linewidth]{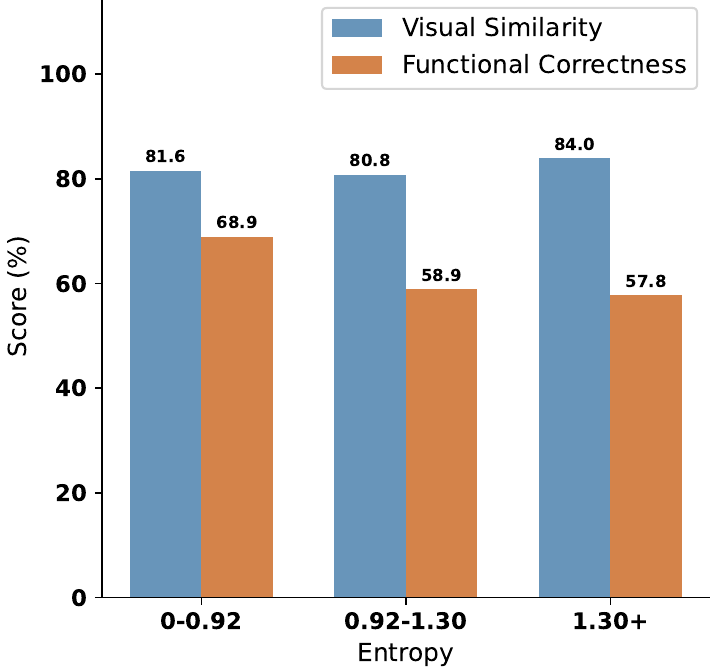}
        \caption{Interaction type}
        \label{fig:analysis_action_type}
    \end{subfigure}
    \hfill
    \begin{subfigure}[t]{0.32\linewidth}
        \centering
        \includegraphics[width=\linewidth]{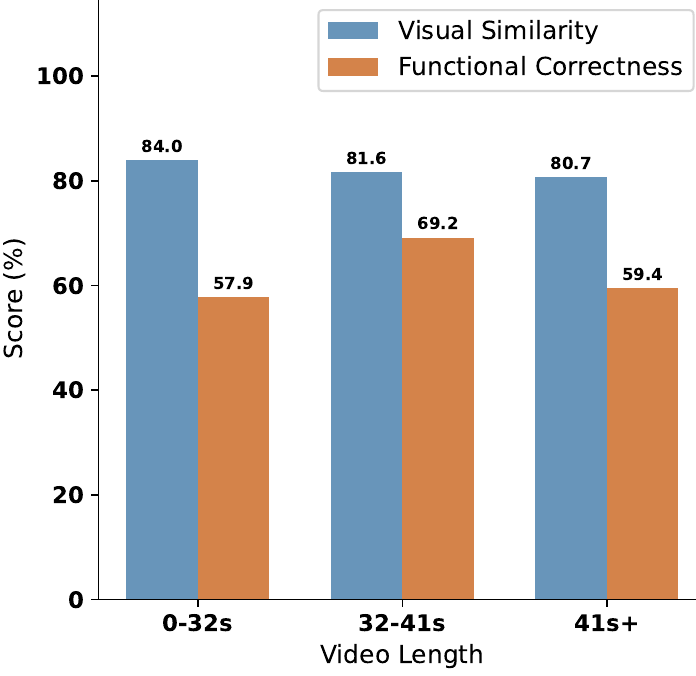}
        \caption{Video length}
        \label{fig:analysis_video_length}
    \end{subfigure}
    \vspace{-2mm}
    \caption{Further analysis of Video2Code on \textit{WebVideo2Code-Real} under different interaction conditions, including interaction count, interaction-type entropy, and video length.}
    \label{fig:further_analysis}
    \vspace{-4mm}
\end{figure*}

\subsection{Further Analysis}

We further analyze Video2Code under different interaction conditions on \textit{WebVideo2Code-Real}, including interaction count, interaction-type diversity, and video length.

\vpara{Effect of Interaction Count.} Figure~\ref{fig:analysis_interaction_count} groups samples by the number of user interactions in each video. The results show that Video2Code performs similarly on videos with 1--8 and 9--16 interactions. Visual similarity remains almost unchanged, while functional correctness shows only a moderate decrease from 62.9 to 59.6. This suggests that Video2Code remains relatively stable as the number of interaction steps increases within the evaluated range.

\begin{table}[t!]
\centering
\small
\renewcommand{\arraystretch}{1.1}
\caption{Effect of frame-based observation on \textit{WebVideo2Code-Real}. For GPT-5.5, we sample frames at 1 FPS and provide them as multi-image context.}
\vspace{-2mm}
\begin{tabular}{lcc}
\toprule
Model 
& \multicolumn{2}{c}{\textit{WebVideo2Code-Real}} \\
\cmidrule(lr){2-3}
& Visual Sim.  
& Func. Corr.    \\
\midrule
GPT-5.5 with 1 FPS Frames & 86.99 & 60.23 \\
% 1 FPS Frames + Claude-Sonnet-4.6 & 87.66 & 68.77 \\
\midrule
Video2Code & \textbf{82.06} & \textbf{61.99} \\
\bottomrule
\end{tabular}
\label{tab:frame_based_baselines}
\vspace{-3mm}
\end{table}

\vpara{Effect of Interaction Type Diversity.} 
% We also analyze the effect of the diversity of interaction types, including clicks, text inputs, selections, and scrolls. Different actions impose different code-generation requirements: clicks and selections often require event handlers and discrete state updates, text inputs require input binding and content-dependent changes, while scrolls require preserving spatial layout and viewport-dependent behavior. 
We measure the diversity of interaction types using Shannon entropy over clicks, text inputs, selections, and scrolls. For each video, let $\mathcal{A}=\{\text{click}, \text{input}, \text{select}, \text{scroll}\}$ denote the set of interaction types, and let $p_a$ be the proportion of interactions of type $a$ in the video. The interaction-type entropy is defined as:
\begin{equation}
H(\mathcal{A}) = - \sum_{a \in \mathcal{A}} p_a \log_2 p_a,
\end{equation}
where interaction types with $p_a=0$ are ignored in the summation. A higher entropy indicates that the video contains a more diverse mixture of interaction types, while a lower entropy indicates that the interaction process is dominated by one or a few action types. As shown in Figure~\ref{fig:analysis_action_type}, visual similarity remains stable across entropy groups, while functional correctness decreases from 68.9 in the low-entropy group to around 58 in higher-entropy groups. This indicates that mixed-action videos are more challenging for executable behavior recovery.

% the two metrics do not present a consistent trend as the interaction-type entropy increases. The visual score stays within a narrow range across the three entropy groups, and the functional score decreases slightly from the low-entropy group to the high-entropy groups but does not show a monotonic or substantial change. This suggests that Video2Code maintains comparable stability on videos with diverse action types and those dominated by fewer action types.

\vpara{Effect of Video Length.} Figure~\ref{fig:analysis_video_length} shows that visual similarity slightly decreases as video length increases, while functional correctness does not show a monotonic drop and is highest in the medium-length group. Overall, Video2Code maintains relatively stable performance across different video lengths, although longer videos may introduce additional temporal grounding challenges.

% Finally, we analyze the effect of video length. Longer videos contain more temporal context and make short action boundaries harder to preserve under limited observation. As shown in Figure~\ref{fig:analysis_video_length}, same as interaction count, video length also has limited effect on both metrics. This suggests that Video2Code maintains relatively stable performance across videos of different lengths.

\subsection{Human Evaluation of Automatic Verification}

To assess the reliability of our browser-based automatic verification, we compare verifier outputs with human judgments on sampled interaction tasks from \textit{WebVideo2Code-Real}. For functional correctness, human annotators inspect the target video segment and generated webpage execution trace to judge whether the demonstrated action is successfully reproduced. The verifier achieves a 96.0\% agreement rate with human judgments. For visual similarity, we compute score differences only on cases where both human and automatic judgments mark the action as successful; cases where either side marks the action as failed are skipped. Under this protocol, 70.2\% of valid cases have an absolute visual-score difference smaller than 10 points, indicating reasonable alignment between automatic and human visual assessment.

\section{Related Work}
\label{gen_inst}

\vpara{Video Understanding with Vision-Language Models.} 
Recent vision-language models have extended multimodal understanding from images to videos, enabling models to process temporal visual inputs and answer questions about dynamic scenes~\cite{cheng2024videollama,chen2024sharegpt4video,li2024llava,zhang2024llava,zhang2025videollama,bai2025qwen2}. 
To handle limited context budgets, video-capable VLMs often rely on sparse frame sampling or compressed visual representations~\cite{cheng2024videollama,chen2024sharegpt4video,zhang2024llava,zhang2025videollama,li2024llama,weng2024longvlm,shen2024longvu}.
Other methods further introduce hierarchical processing or adaptive frame/clip selection to preserve informative temporal regions~\cite{cheng2025scaling,hannan2025revisionllm,shu2025video,tang2025adaptive,tan2026think}.
These strategies are effective for many natural video understanding tasks, where the goal is often to capture high-level semantic content over time. 
In contrast, UI interaction videos contain temporally sparse but functionally decisive events: a short click, selection, or scroll may determine the executable behavior that must be implemented in code.
Thus, Video2Code focuses not on general video understanding, but on recovering action-conditioned UI state transitions for executable webpage generation.

\vpara{Visual UI-to-Code Generation.} Generating front-end code from visual inputs has attracted increasing attention, from early screenshot-to-code systems such as pix2code~\cite{beltramelli2018pix2code} to recent VLM-based benchmarks and datasets, including WebSight~\cite{laurencon2024websight}, Vision2UI~\cite{gui2024vision2ui}, Design2Code~\cite{si2024design2Code}, Web2Code~\cite{yun2024web2code}, and WebCode2M~\cite{gui2025webcode2m}. 
Most existing methods focus on reconstructing webpage layout, style, and visual appearance from static screenshots or design images, with recent improvements from decomposition, hierarchical generation, or modular agentic pipelines~\cite{wan2024automatically,gui2025uicopilot,jiang2025screencoder}. 
Beyond one-shot static reconstruction, UI2Code$^{\text{N}}$ formulates UI-to-code as iterative visual optimization~\cite{yang2025ui2code}, while WebVIA studies interactive and verifiable UI-to-code generation from multi-state interaction evidence~\cite{xu2025webvia}. 
Video2Code complements these efforts by using UI interaction videos as input, where user actions and state transitions provide temporal evidence for preserving both visual appearance and executable behavior.

\section{Conclusion}

In this paper, we propose \textbf{Video2Code}, an action-aware approach for generating executable webpages from UI interaction videos. We formulate UI video-to-code generation as executable state-transition recovery and introduce \textit{Action-Aware Revisit}, which identifies action-critical regions from coarse video observation and revisits them at higher temporal resolution. With action-aligned video-code supervision, Video2Code learns to generate code from both global video context and local transition evidence. Experiments under visual and functional evaluation show that Video2Code improves behavior-faithful UI video-to-code generation over direct video observation.

% \clearpage

\section*{Limitations}

Video2Code focuses on generating executable webpages from UI interaction videos, but its current scope does not cover the full diversity of modern web applications. While \textit{WebVideo2Code-Bench} includes both real-world and synthetic webpages, it mainly focuses on publicly accessible pages and does not fully cover scenarios involving authentication, personalization, backend-dependent content, or highly asynchronous behavior. Our browser-based functional verification focuses on common webpage operations, including clicks, text entry, selection, and scroll. Although these operations cover many frequent UI behaviors, they do not yet include more complex interactions such as drag-and-drop, file upload, canvas manipulation, or long-horizon workflows involving external services. Expanding the benchmark and verifier to support richer interaction primitives would enable a more comprehensive evaluation of behavior-faithful UI video-to-code generation.

\section*{Ethical Statement}

This work studies UI video-to-code generation for browser-interactable webpages. We recognize potential concerns related to privacy, copyright, and misuse, especially when real-world webpages are recorded or when generated code is used to imitate existing interfaces. Video2Code is intended as a research framework for studying behavior-faithful webpage generation, rather than for cloning deployed services or collecting private user data.

For data construction and evaluation, we use controlled synthetic webpages and publicly accessible real-world webpages. We do not use authentication-required, paywalled, or private pages. Real-world recordings are limited to common and reversible interactions such as clicks, text entry, selections, and scrolls, and examples with unstable or ambiguous behavior are filtered. We do not intentionally collect personal information, credentials, or private user-generated content.

We use AI-assisted annotation during supervision construction, and all generated supervision is filtered before training. Human annotation is limited to action timestamp labeling and verification reliability assessment, without collecting sensitive annotator information. Released artifacts will include documentation describing intended research use, construction procedures, and limitations.

\bibliography{references}

\appendix

\clearpage

\section{Appendix}

% 1. 数据构建：1）webvia-video-agent与原本的详细区别，如何实现提取timeline的 2） process数据流程细节，从agent探索提取出来的视频怎么被构建成问答对

% 2. 训练：1）数据总量，以及内容分布（多少html,多少url） 2) 训练细节（抽帧策略）

% 3. benchmark & 测试细节

\subsection{Data Construction}
\label{app:data_construction}

We provide additional implementation details for constructing \textit{WebVidCoding}, the action-aligned video-code dataset used to train Video2Code. Specifically, we describe browser-based webpage exploration, action timeline extraction, clipping window construction, and two-round video-code supervision generation.

\vpara{Exploration protocol.} We use an automated browser exploration agent to collect UI interaction videos from synthetic HTML documents and real-world URLs. Given a webpage URL or an HTML file, the agent first identifies candidate interactive elements from the rendered page and DOM tree, and then explores the page by executing common operations, including clicks, text entry, selections, and scrolls. The exploration is performed over webpage states rather than isolated elements: after each operation, the browser waits for the page to stabilize and records the resulting viewport state. We remove repeated operations that lead to visually identical or near-identical states, so that the collected trajectory focuses on distinct action-triggered transitions. During exploration, the browser viewport is recorded as a video, and the executed action sequence is stored together with action metadata.

\vpara{Action-boundary localization.} To localize user actions in the recorded video, we inject a lightweight visual marker during recording. The marker is a 12-pixel-high bar fixed at the bottom of the viewport and changes color when an action is executed. After recording, we detect color changes in the marker region and use them as temporal anchors for user operations. Each anchor is associated with the corresponding action type, such as click, text entry, selection, or scroll.

\vpara{Action-critical clipping.} Given an action anchor, we construct a clipping window by expanding the anchor with pre-action and post-action temporal margins. The pre-action margin preserves the UI state before the operation, while the post-action margin captures the resulting state after the page response. For actions whose effects are only visible after a subsequent scroll, we merge the triggering action and the required scroll into a single action-critical segment, so that the clipped segment preserves complete transition evidence.

\vpara{Two-round supervision generation.}
We construct the two-round supervision data through an offline teacher-generation pipeline. 
For each training sample, we first use the action-aligned timeline to derive action-critical clipping windows, and sample the corresponding clips into frame sequences for teacher annotation. These frames are provided to Gemini-3.1-Pro to generate supervision targets for the two-round Action-Aware Revisit format.
In the first round, the target output is a structured tool-call trajectory that invokes \texttt{extract\_multi\_segment} with the action-critical temporal windows. This teaches the model to produce revisit plans that are consistent with the inference-time clipping interface. 
In the second round, the returned clips are appended to the model context, and the target output consists of an interaction-aware observation and reasoning process enclosed by \texttt{<think></think>}, followed by the executable webpage implementation in HTML, CSS, and JavaScript.
To improve supervision quality, we generate the second-round target with two separate teacher requests: one for visual observation and interaction reasoning, and the other for executable code generation. 
This separation reduces interference between high-level transition analysis and low-level code synthesis, producing cleaner supervision for both revisit-based reasoning and final webpage reconstruction.

% \yz{YZ: refine it by zhen}
% \yz{Do we need to add more detail here?}

\vpara{Quality filtering.} We apply filtering to remove examples with unreliable supervision. 
We discard examples with invalid tool-call format, invalid thinking format or incomplete codes. We also filter out examples where the target implementation fails to render into a valid webpage. 
The remaining examples form \textit{WebVidCoding}, a 12K-example dataset consisting of 4K synthetic webpage examples and 8K real-world webpage examples.

\subsection{Training Details}\label{app:training_details}

We provide implementation details of the supervised fine-tuning procedure used to train Video2Code. 

\vpara{Training format.} Each training instance is constructed from an interaction video $v$, an action-aligned clipping timeline $\tau$, the returned clips $E(v,\tau)$, and the corresponding executable webpage implementation $y$. The first round takes the full interaction video as input and supervises the model to predict a structured temporal clipping call:
\[
\texttt{extract\_multi\_segment}(v, \tau).
\]
The target segments in $\tau$ are derived from the action timeline and are expected to cover action-critical state transitions. The second round appends the returned clips to the model context and supervises the model to generate the final HTML/CSS/JavaScript implementation.

\vpara{Tool-call target format.} The tool-call target contains a list of temporal intervals, each represented by its start and end timestamps. Each interval is constructed to cover the pre-action state, the action execution process, and the post-action state. For action-scroll cases, a single interval may cover both the triggering action and the subsequent scroll when both are needed to observe the resulting state.

\vpara{Code-generation target.} The code-generation target is a self-contained webpage implementation. The output is required to include the HTML structure, CSS styling, and JavaScript logic needed to reproduce the visual states and action-triggered behaviors demonstrated in the video. This format ensures that the model learns to generate executable code from both global video context and local transition evidence recovered through revisit.

\vpara{Training configuration.} We fine-tune Video2Code for 3 epochs with a learning rate of $2\times10^{-5}$ and a batch size of 32 on H800 GPUs. Table~\ref{tab:training_details} summarizes the training configuration.

\begin{table}[t]
\centering
% \small
\caption{Training configuration of Video2Code.}
\setlength{\tabcolsep}{2.0pt}
\renewcommand{\arraystretch}{1.1}
\begin{tabular}{lc}
\toprule
Configuration & Value \\
\midrule
Base model & GLM-4.6V-100B \\
Training dataset & \textit{WebVidCoding} \\
Number of examples & 12K \\
Training objective & Supervised fine-tuning \\
Input format & Full video + returned clips \\
Tool-call target & \texttt{extract\_multi\_segment} \\
Code target & HTML/CSS/JavaScript \\
Frames per clip & 3 fps \\
Learning rate & $2\times10^{-5}$  \\
Batch size & 32 \\
Epochs & 3 \\
Hardware & H800 GPUs \\
\bottomrule
\end{tabular}
\label{tab:training_details}
\vspace{-3mm}
\end{table}

\subsection{Benchmark Construction}\label{app:benchmark_contruction}

We provide additional details for constructing \textit{WebVideo2Code-Bench}. 
The benchmark is designed to evaluate whether a generated webpage can reproduce both the visual states and the action-triggered behaviors demonstrated in a UI interaction video. 
It contains two subsets, \textit{WebVideo2Code-Real} and \textit{WebVideo2Code-Syn}, covering real-world webpages and controlled synthetic webpages respectively.

\vpara{\textit{WebVideo2Code-Real}.} We construct the real-world subset from manually recorded webpage interaction videos. We start from 100 publicly accessible websites covering diverse real-world webpage scenarios, including \textit{education platforms}, \textit{academic, legal, and database-style websites}, \textit{health information and symptom-checking pages}, \textit{finance calculators and report interfaces}, \textit{shopping and e-commerce flows}, \textit{news search}, \textit{online tools such as translation and file conversion},  \textit{job search platforms}, and \textit{travel booking or itinerary search pages}. For each website, we operate the webpage in a browser and record a UI interaction video that contains common user operations, including clicks, text entry, selections, and scrolls. We manually annotate the timestamps of user actions and use them to segment the video into task-wise state transitions. We filter out recordings with overly sparse interactions, overly dense interactions, unstable page behavior, ambiguous action effects, or interaction processes that cannot be reliably verified. After filtering, \textit{WebVideo2Code-Real} contains 85 real-world webpage examples.

\vpara{\textit{WebVideo2Code-Syn}.} We construct the synthetic subset from executable webpages explored by the browser exploration agent. The agent records interaction trajectories during webpage exploration, and we further convert these trajectories into concise benchmark videos. Specifically, we remove repeated operations, redundant states, and transitions whose effects are visually ambiguous or difficult to verify. Compared with the real-world subset, the synthetic subset provides more controlled interaction structures and executable references, which allows us to evaluate UI video-to-code generation under cleaner webpage behavior. After filtering, \textit{WebVideo2Code-Syn} contains 76 synthetic webpage examples.

\vpara{Task segmentation.} Each benchmark video is decomposed into multiple interaction tasks according to the annotated action timestamps. A task corresponds to one user operation and its surrounding state transition. For each task, we keep frames covering the pre-action state, the action execution process, and the post-action state. These task-wise frame sequences are used by the browser-based verifier to select the corresponding operation on the generated webpage and judge whether the resulting execution trace matches the target transition.

\vpara{Filtering criteria.} We apply filtering to ensure that each benchmark example is clear and evaluable. Examples are removed if the webpage fails to load, the recording is incomplete, the interaction effect is not visually observable, the action timestamp is ambiguous, or the task requires unsupported operations beyond the verifier action space. For synthetic webpages, we additionally remove trajectories with repeated or redundant operations that do not introduce new UI states. This filtering process ensures that the benchmark focuses on observable and executable webpage behavior.

\subsection{Evaluation Protocol and Metrics}\label{app:evaluation_metrics}

We provide additional details of the browser-based evaluation protocol used in \textit{WebVideo2Code-Bench}. Given a UI interaction video and a generated webpage implementation, the goal is to evaluate whether the generated webpage can reproduce both the visual states and the action-triggered behaviors demonstrated in the video.

\vpara{Task-wise video segmentation.} We first split each input interaction video into task-wise frame sequences according to the annotated user-action timestamps. Each task corresponds to one user operation and its surrounding state transition. For example, an input video can be decomposed into multiple tasks:
\[
v \rightarrow \{T_1, T_2, \ldots, T_n\},
\]
where each task $T_i$ contains frames covering the pre-action state, the user operation, and the post-action state. These task-wise frames serve as the target reference for both action execution and result verification.

\vpara{Generated webpage rendering.} For each generated implementation, we launch a browser environment and render the generated webpage. Before executing each task, we record the current webpage screenshot and extract the DOM tree. The screenshot provides visual context, while the DOM tree provides structural information about interactive elements on the generated page.

\vpara{Action selection and execution.} For each task, the verifier receives the current webpage screenshot, the DOM tree, and the target task frames from the input video. It then selects the operation that best matches the demonstrated user action and executes it on the generated webpage. The supported action space includes click, text entry, selection, and scroll. During execution, the browser records the resulting webpage trace as a sequence of frames.

\vpara{Task-level scoring.} After executing each task, the verifier compares the recorded webpage trace with the target task frames and produces two task-level scores. (1) \textit{Task-level functional correctness.} Functional correctness evaluates whether the generated webpage produces the expected state change after the demonstrated user action is executed. For each task, the verifier checks whether the operation can be successfully performed and whether the resulting webpage state matches the transition shown in the target video segment. The task receives a binary functional score of $1$ if the expected state change occurs, and $0$ otherwise. If the operation cannot be executed, the webpage fails to respond, or the resulting state is inconsistent with the target transition, the task is counted as functionally incorrect. After all tasks of a webpage have been verified, the page-level functional score is reported on a 100-point scale by multiplying the ratio of functionally correct tasks among all verified tasks by $100$. In other words, it is computed as the number of tasks with a functional score of $1$ divided by the total number of tasks for that webpage, and then multiplied by $100$. (2) \textit{Task-level visual similarity.} Visual similarity evaluates whether the resulting webpage state after execution visually matches the target post-action state. For each functionally correct task, the verifier compares the recorded execution trace of the generated webpage with the target task frames and assigns a visual similarity score on a 100-point scale. Failed functional executions are excluded from page-level visual similarity, since the generated webpage does not reach the intended post-action state and comparing it with the target state would not reflect the quality of the intended visual reconstruction. The page-level visual similarity is therefore computed as the average task-level visual similarity over all functionally correct tasks.

To ensure that visual verification also covers the initial page state, we introduce a separate initial-page comparison mechanism. For long webpages, this state is naturally covered by the initial scroll task: we first perform a scroll action to reveal the full initial page beyond the viewport, thereby verifying both the page's scrollability and its visual similarity to the original initial-page view. However, for smaller webpages whose initial page is fully contained within the viewport, no scroll task is needed. In such cases, the first task may directly evaluate the post-action state after a click, text entry, or selection, leaving the initial page without a corresponding visual similarity score. Therefore, we conduct an additional visual comparison between the rendered initial page and the first frame of the webpage video, and include the resulting score in page-level similarity.

% \yz{refine it by zhen}

\vpara{Sequential execution.} For \textit{WebVideo2Code-Real}, tasks are executed sequentially following the order of the recorded interaction video. After one task is executed, the browser state is preserved and used as the starting state for the next task. This setting matches the real recording process, where user operations are performed continuously on the same webpage session. For \textit{WebVideo2Code-Syn}, we follow the predefined task structure of the synthetic trajectory. When a task is annotated as requiring a fresh initial state, we reload the generated webpage before executing the next task; otherwise, we continue from the current browser state. This handles synthetic trajectories where some operations are designed to be evaluated independently from the initial page.

\vpara{Initial-page retry for \textit{WebVideo2Code-Real}.} To reduce cascading failures, we use an initial-page retry mechanism. When a task cannot be executed from the current generated webpage state, we reload the generated webpage and retry the same task from the initial page. If the retry succeeds, the new browser state is used for subsequent tasks; otherwise, the task is marked as functionally incorrect. This retry mechanism reduces cascading failures, where one failed operation would otherwise cause later independently executable tasks to be skipped. 
% \yz{only used for real world bench}

\vpara{Page-wise aggregation.} Both metrics are first computed at the task level and then aggregated at the page level. For a webpage $p$ with task set $\mathcal{T}_p$, we compute
\[
\small
\begin{aligned}
\mathrm{Vis}(p) &= \frac{1}{|\mathcal{T}_p|}
\sum_{t\in\mathcal{T}_p} v_{p,t}, \\
\mathrm{Func}(p) &= 
\frac{\left|\{t \in \mathcal{T}_p \mid f_{p,t}=1\}\right|}
{|\mathcal{T}_p|} \times 100 .
\end{aligned}
\]
where $v_{p,t}$ denotes the visual similarity score for task $t$, and $f_{p,t}\in\{0,1\}$ denotes whether task $t$ is functionally correct. Thus, $\mathrm{Func}(p)$ measures the percentage of successfully reproduced tasks on webpage $p$.

The final benchmark score is obtained by averaging webpage-level scores over all webpages:
\[
\small
\begin{aligned}
\mathrm{Vis} &= \frac{1}{|\mathcal{P}|}
\sum_{p\in\mathcal{P}} \mathrm{Vis}(p), \\
\mathrm{Func} &= \frac{1}{|\mathcal{P}|}
\sum_{p\in\mathcal{P}} \mathrm{Func}(p).
\end{aligned}
\]
This page-wise aggregation prevents webpages with more interaction tasks from dominating the overall result.

% \yz{refine by zhen}

\subsection{Prompts}

We provide the prompt templates used in our data construction, model inference, and browser-based evaluation. These prompts cover the main stages of Video2Code, including synthetic webpage construction, temporal clipping tool-call generation, video-to-code supervision, baseline inference, action replay, and interaction verification. 

% \yz{add all prompts -- reine by zhen}

\vpara{Synthetic HTML document construction prompts.} We use synthetic webpages in both data construction and evaluation. Each synthetic HTML document is constructed through a two-stage process. First, we prompt a teacher model to generate a webpage design instruction based on a given theme. Second, we provide the generated instruction to a code-generation prompt to produce an executable HTML document. This process increases the content and interaction diversity of the synthetic webpages. The complete prompt templates are shown in Figure~\ref{fig:webpage-design-template} and Figure~\ref{fig:code_generate_prompt}.

\vpara{Tool-call segment generation prompt.} During data construction, we construct the first round supervision by a given UI interaction video frames and its action timeline. The prompt asks the annotator/model to identify temporal regions that contain complete state-action-state evidence based on timeline and output structured calls to \texttt{extract\_multi\_segment}, where each argument specifies an action-critical temporal interval to revisit. The complete prompt is shown in Figure~\ref{fig:segment-template}.

\vpara{Video-to-code generation prompt.} During the second-round supervision, we use two prompts to separate reasoning from implementation. The first prompt requires the model to analyze the extracted video clips and frames and produce its reconstruction reasoning enclosed within \texttt{<think>} and \texttt{</think>} tags. The second prompt then asks the model to generate the final executable webpage code in HTML, CSS, and JavaScript. The generation prompt emphasizes both visual fidelity and interactive functionality, including event handlers, state updates, input binding, selections, and scroll-dependent behavior. The complete prompts are shown in Figure~\ref{fig:thinking-template} and Figure~\ref{fig:generation-template}.

\vpara{Video-based baseline inference prompt.} This prompt is used for closed-source and open-source VLM baselines that support native video input. Unlike Video2Code, these baselines are prompted to directly generate executable webpage code from the full UI interaction video, without tool-based temporal revisit. The complete prompt is shown in Figure~\ref{fig:prompt_video_baseline}.

\vpara{Frame-based baseline inference prompt.} This prompt is used for baselines evaluated under frame-based observation. We convert the input video into sampled frames and provide them as multi-image context. The model is then prompted to directly generate executable HTML/CSS/JavaScript code from these frames, using the same task objective as the video-based setting. The complete prompt is shown in Figure~\ref{fig:prompt_frame_baseline}.

\vpara{Interaction selection prompt for evaluation.} During browser-based evaluation, each task is evaluated by selecting and executing the operation that best matches the demonstrated user action. This prompt asks the model to choose the appropriate action based on the target video frames, the current rendered webpage screenshot, and the page DOM tree. The complete prompt template is shown in Figure~\ref{fig:action_replay_prompt}.

\vpara{Interaction verification prompts for evaluation.} We use two interaction verification prompts. The first prompt evaluates the generated webpage after the selected action is executed. It judges whether the webpage produces the expected state change, and assigns a visual similarity score for the resulting state against the target post-action state. The second prompt is used for initial-page visual verification, where the rendered initial page is compared with the first frame of the input video when the initial page is not otherwise covered by a task-level evaluation. The complete prompts are shown in Figure~\ref{fig:visual-similarity-template} and Figure~\ref{fig:initial-state-visual-similarity-template}.

\begin{figure}[t]
\centering
\begin{tcolorbox}[
    colback=gray!2,
    colframe=blue!70!black,
    colbacktitle=blue!12!white,
    title=\textbf{Synthetic HTML Design Instruction Template},
    fonttitle=\bfseries,
    coltitle=blue!50!black,
    enhanced,
    sharp corners,
    boxrule=0.5pt,
    left=5pt,
    right=5pt,
    top=5pt,
    bottom=5pt,
    titlerule=0mm,
    width=0.5\textwidth,
    title style={left color=blue!8!white, right color=blue!5!white},
]
\small
Please write a detailed prompt that will be used to instruct a text-to-image model to generate an interactive webpage HTML code with the theme \textbf{``<INSERT THEME FROM THEME LIST>''}.

\vspace{2mm}

\textbf{Specific Requirements:}
\begin{enumerate}
    \item The webpage content should revolve around the specified theme and include a wide variety of theme-related modules.
    \item The webpage must contain multiple interactive elements, limited to \texttt{buttons}, \texttt{input fields}, and \texttt{dropdown selectors}. Each interactive component should cause corresponding and reasonable changes on the webpage.
    \item The webpage content should be rich, detailed, and contextually diverse.
    \item The output should only contain the final prompt for the AI to generate the webpage—without explanations, metadata, or additional commentary.
\end{enumerate}
\end{tcolorbox}
\caption{Template used to construct webpage design prompts for generating interactive webpage HTML code.}
\label{fig:webpage-design-template}
\end{figure}

\begin{figure*}[t!]
\centering
\begin{tcolorbox}[
    colback=gray!2,
    colframe=blue!70!black,
    colbacktitle=blue!12!white,
    title=\textbf{Synthetic HTML Code Generation Prompt},
    fonttitle=\bfseries,
    coltitle=blue!50!black,
    enhanced,
    sharp corners,
    boxrule=0.5pt,
    left=6pt,
    right=6pt,
    top=5pt,
    bottom=5pt,
    titlerule=0mm,
    width=0.98\textwidth, % ✅ 跨双栏
    title style={left color=blue!8!white, right color=blue!5!white}
]
\small
You are a web development expert highly sensitive to details and interaction experience, proficient in React and Tailwind CSS. 
Please generate a highly interactive single-page application with reasonable layout and rich content for the specified theme according to the following requirements.

\vspace{2mm}
\textbf{Basic Requirements:}
\begin{enumerate}
    \item Generate a complete interactive single-page website rendered using \textbf{React (v18)} and \textbf{Tailwind CSS (v3+)}. 
    \item Return only the full source code wrapped within \texttt{<html>...</html>} tags. \textbf{Do not} include markdown wrappers, explanations, or code comments.
    \item Must include the following dependencies:
\begin{lstlisting}[basicstyle=\ttfamily\footnotesize, frame=none, breaklines=true]
<script src="https://cdn.jsdelivr.net/npm/react@18.0.0/umd/react.development.js"></script>
<script src="https://cdn.jsdelivr.net/npm/react-dom@18.0.0/umd/react-dom.development.js"></script>
<script src="https://cdn.jsdelivr.net/npm/@babel/standalone/babel.js"></script>
<script src="https://cdn.tailwindcss.com"></script>
<link rel="stylesheet" href="https://cdnjs.cloudflare.com/ajax/libs/font-awesome/5.15.3/css/all.min.css"></link>
\end{lstlisting}
\end{enumerate}

\vspace{-1mm}
\textbf{Interactivity and Functional Areas:}
\begin{enumerate}
    \item All interactive components (\texttt{input}, \texttt{button}, \texttt{select}) must trigger meaningful updates to the rendered page.
    \item For editable content, use modals, dropdowns, or input forms with complete validation.
    \item Use real pictures from \url{https://picsum.photos/}. Each image must have a fixed URL and remain constant across reloads.
\end{enumerate}

\vspace{-1mm}
\textbf{Page Structure and Layout:}
\begin{enumerate}
    \item Include logical partitions (navigation, sidebar, main content, etc.) referencing modern app layouts.
    \item Ensure all sections are populated; empty placeholders are not allowed.
    \item The visual style must match the assigned theme (e.g., business, minimalism, tech, lifestyle).
\end{enumerate}

\vspace{-1mm}
\textbf{Notes:}
\begin{itemize}
    \item Do not output explanations or text outside the code.
    \item Ensure all theme-related UI logic is complete and intuitive.
\end{itemize}

\vspace{1mm}
\textbf{Webpage Description:} \texttt{<INSERT DETAILED PROMPT FROM STAGE 1>}
\end{tcolorbox}
\caption{Code Generation Prompt used for large-scale HTML synthesis.}
\label{fig:code_generate_prompt}
\end{figure*}
\begin{figure*}[t]
\centering
\begin{tcolorbox}[
    colback=gray!2,
    colframe=blue!70!black,
    colbacktitle=blue!12!white,
    title=\textbf{Tool-Call Segment Generation Prompt for Data Construction},
    fonttitle=\bfseries,
    coltitle=blue!50!black,
    enhanced,
    sharp corners,
    boxrule=0.5pt,
    left=5pt,
    right=5pt,
    top=5pt,
    bottom=5pt,
    titlerule=0mm,
    width=\textwidth,
    title style={left color=blue!8!white, right color=blue!5!white},
]
\small
You are a video understanding expert proficient in user interface (UI) interaction analysis, and you also have a deep background in frontend engineering.

\vspace{1mm}

\textbf{Task Background}

I will provide you with a \textbf{URL of a long video file}, as well as a set of \textbf{short video screenshots (image sequences) that I have prepared for you}.
These screenshots are key frames from the moments when the user performs key interaction operations, such as clicking, typing, selecting, and scrolling, in the long video, covering the initial state before the operation, during the operation, and the page changes after the operation. The user will also give you an approximate description of what operation was performed in the long video for each group of operations.
This video URL: \textbf{<VIDEO\_URL>}

\vspace{1mm}

\textbf{Core Task}

You need to map the operation list I send you to these short video screenshots in your mind, and then \textbf{fabricate a thinking process in which you are watching the complete long video}. In this process, you should use the visual page content, that is, the screenshots you refer to, to \textbf{describe in what time period each interaction occurred, what exactly was done, and what feedback the page produced}.
Finally, you need to call a tool to batch-cut these segments.

\vspace{1mm}

\textbf{Output Structure Specification}

Please strictly follow the following requirements in your output, and do not include any greetings:

\vspace{1mm}

\textbf{Output Requirement 1: Long Video Observation and Fabricated Thinking Process \texttt{<think></think>}}

This is the key point. Please complete the above ``pretending to independently discover and observe'' process inside the \texttt{<think></think>} tags, pretending that you are studying the interaction time points in the webpage, and describe them in detail.
You must strictly follow the following rules:

\begin{enumerate}
    \item \textbf{Pretend to discover independently}: Do not mention words such as ``reference operation list,'' ``user-provided list,'' or ``which image'' in the thinking process. Your tone must sound like you are watching the complete video frame by frame and then independently discovering these operations.
    
    Incorrect example: ``According to reference operation 1, I see...''
    
    Correct example: ``While browsing the beginning of the video, I notice that around 4.5 seconds, the mouse cursor moves to the `Login' button, and then the page jumps...''
    
    \item \textbf{Describe the operation and visual feedback in detail}: For each time segment you mention, you may slightly modify the times based on the times in the following \textbf{user timeline and operation reference description} to make it look more like an independent discovery. You must precisely describe what happens on the screen during that period. This includes:
    \begin{itemize}
        \item What specific button or link did the mouse click?
        \item What text was entered into the input field?
        \item How did the page respond to this, such as popup, color change, or page jump?
        \item Use the visual information from these short video screenshots, such as which one is before, which one is after, or the operation process, to enrich your description.
    \end{itemize}
    
    For example: ``At 4.5 seconds, I see an initial page layout, then the mouse moves and clicks, and at 5.0 seconds the page finishes loading.''
    
    \item \textbf{Special treatment}: For operations marked as \textbf{scroll browsing}, there is no need to describe the specific operation details; you only need to describe that scrolling shows the overall structure of the entire page.
\end{enumerate}

\vspace{1mm}

\textbf{Output Requirement 2: Tool Call Statement and Tool Call \texttt{<tool></tool>}}

After completing the thinking, briefly state in natural language that you have identified the key interaction segments and will call the tool to cut and extract them.

Example: After analysis, I have independently identified the key interaction operations in the video. I will call the \texttt{zai\_extract\_multi\_video\_segments} tool to precisely cut and extract the following key interaction segments: from 3.5 seconds to 5.5 seconds, from 11.2 seconds to 13.2 seconds.

\vspace{1mm}

\textbf{Output Requirement 3: Tool Call \texttt{<tool></tool>}}

Finally, inside the \texttt{<tool></tool>} tags, output JSON data that strictly conforms to the following format.

Note: The tool accepts two main parameters: \texttt{video\_url}, which is the URL I pass to you, and \texttt{segments}, which is a list of objects containing \texttt{start} and \texttt{end}. The \texttt{video\_url} part must be written as the exact same URL I pass to you.

JSON format example:

\{ ``video\_url'': ``\textbf{<VIDEO\_URL>}'', ``segments'': [\{ ``start'': 3.5, ``end'': 5.5 \}, \{ ``start'': 11.2, ``end'': 13.2 \}] \}

\vspace{1mm}

\textbf{Timeline information}

\textbf{<TIMELINE AND OPERATION INFORMATION>}

\vspace{1mm}

Please strictly follow the above output structure specification and do not output any irrelevant information.

\end{tcolorbox}
\caption{Prompt for generating tool-call segments from interaction videos.}
\label{fig:segment-template}
\end{figure*}

\begin{figure*}[t]
\centering
\begin{tcolorbox}[
    colback=gray!2,
    colframe=blue!70!black,
    colbacktitle=blue!12!white,
    title=\textbf{Video-to-Code Thinking Reconstruction Prompt for Data Construction},
    fonttitle=\bfseries,
    coltitle=blue!50!black,
    enhanced,
    sharp corners,
    boxrule=0.5pt,
    left=5pt,
    right=5pt,
    top=5pt,
    bottom=5pt,
    titlerule=0mm,
    width=\textwidth,
    title style={left color=blue!8!white, right color=blue!5!white},
]
\small
You are a senior engineer proficient in frontend development (React + Tailwind).

\vspace{1mm}

\textbf{Task Objective}

Based on the \textbf{key short videos capturing the exact moments when interactions occur} that you observed in the previous round from a long webpage interaction video, you need to construct a \textbf{visualized video observation process}, while also \textbf{pretending that you are about to generate HTML code to replicate this interactive webpage}.

\vspace{1mm}

\textbf{Core Task}

Your task is to ``reverse-render'' this group of \textbf{operation lists} in your mind into a \textbf{visualized video observation process}.

\vspace{1mm}

\textbf{<VIDEO FRAMES INFORMATION>}

\vspace{1mm}

\textbf{Output Structure Specification}

Please strictly follow the requirements below when outputting, and do not include any irrelevant greetings or extra text.

\vspace{1mm}

\textbf{Output Requirement: Interaction Logic Observation Flow \texttt{<think></think>}}

\textbf{This is the key point!} Please complete the thinking process of observing each short video inside the \texttt{<think></think>} tags, pretending that you are studying the interaction points on the webpage and trying to replicate them.

You need to strictly follow the rules below:

\begin{enumerate}
    \item \textbf{First observe the overall webpage layout}: Before watching the first interaction short video, first inspect the webpage layout in the first video. This is the initial layout of the webpage, and you should describe it in detail.

    \item \textbf{Observe each operation}:
    \begin{itemize}
        \item \textit{Observe the mouse position}: For each interaction short video, please observe the \textbf{changes in the mouse position}, because it must be the mouse position interacting with some interactive element that causes the page to change.
        \item \textit{Confirm the clicked object}: By observing the feedback after the click, clearly confirm in your thinking exactly which element the mouse clicked.
        \item \textit{Construct the interaction narrative}: Combined with properties such as ``scrolling after the operation'' and ``continuous interaction'' in the prompt, describe the page state changes caused by the operation, for example: loading spinner, page navigation, adding a new row of data to a table, scrolling to reveal content, etc.
    \end{itemize}

    \item[\textbf{2.5}] \textbf{Special treatment for scroll browsing}: When the operation property indicates that this operation is a \textbf{scroll browsing} operation, there is no need to describe the above details. You only need to piece together the screenshots in the scroll browsing process and describe the overall structure of the entire page.

    \item \textbf{Summarize and pretend that you are about to write code}: At the end, summarize what functions you have observed that this webpage needs to have, and state that you are starting to write the code. You do not actually need to write it.
\end{enumerate}

\vspace{1mm}

\textbf{Important Constraints}

\begin{itemize}
    \item At this stage, do not mention any code terms, such as \texttt{div}, \texttt{state}, \texttt{onclick}, etc.; only describe the visuals and logic.
    \item You are not looking at images; you are watching videos. Do not output anything inside \texttt{<think></think>} suggesting that you are ``pretending to watch videos'' or reading ``Figure 1''. Instead, fully fabricate it: only say that you are watching these clipped videos yourself.
    \item Do not mention any video timestamps; only describe the video content.
    \item Do not mention the reference operation information sent by the user. Instead, disguise it as something you observed from the videos yourself. That is, do not directly output ``Operation 1'' or similar reference content provided by the user in the result.
\end{itemize}

\end{tcolorbox}
\caption{Prompt used to guide the model to produce an interaction logic thinking flow from clipped webpage interaction video frames before code generation.}
\label{fig:thinking-template}
\end{figure*}

\begin{figure*}[t]
\centering
\begin{tcolorbox}[
    colback=gray!2,
    colframe=blue!70!black,
    colbacktitle=blue!12!white,
    title=\textbf{Video-to-Code Coding Reconstruction Prompt for Data Construction},
    fonttitle=\bfseries,
    coltitle=blue!50!black,
    enhanced,
    sharp corners,
    boxrule=0.5pt,
    left=5pt,
    right=5pt,
    top=5pt,
    bottom=5pt,
    titlerule=0mm,
    width=\textwidth,
    title style={left color=blue!8!white, right color=blue!5!white},
]
\small
You are highly skilled at building interactive web pages with React and Tailwind, and can precisely restore a complete HTML interactive webpage based on multiple webpage screenshots provided by the user.

\vspace{1mm}

\textbf{Interface Requirements}
\begin{enumerate}
    \item Build the initial page according to the first webpage screenshot provided by the user, and it must be completely consistent with the content of the first screenshot given to you.
    \item Do not omit any details, including background colors, fonts, font sizes, spacing, borders, icons, text, etc., all of which must strictly match the screenshots.
    \item Every sentence of text in the screenshots must be presented exactly as it is.
    \item For image content, please use real images from the \url{https://picsum.photos/} library, with URLs similar to \url{https://picsum.photos/id/.../.../...}. Each image must explicitly list its URL; do not use reusable image components. The image URL of each webpage component must be fixed, and do not use random numbers to regenerate it each time.
\end{enumerate}

\vspace{1mm}

\textbf{Core Task Requirements}

Please bring out a ``microscope''-level power of observation to achieve a 1:1 pixel-level reproduction of the webpage interactions:

\begin{enumerate}
    \item \textbf{Observe the overall page changes}: Carefully observe the timeline screenshot sequence provided below. Focus on comparing the \textbf{initial screen before the operation} and the \textbf{final state/instant-change state}, and do not omit any details on the page, such as a popup notification appearing in the upper-right corner, a newly added list item, etc. For screenshot sequences marked as \textbf{pure scroll browsing}, treat them as structured puzzle pieces for obtaining the global layout and hidden content of the page, ensuring the completeness of the entire page.
    
    \item \textbf{Precisely locate the source of interaction}: In the \textbf{initial screen before the operation} screenshot, find the mouse position, infer which component was clicked or typed into, and ensure that these components exist in the reproduced webpage and possess the same capabilities as the original. Make its functionality and visual feedback exactly the same as in the screenshots.
    
    \item All interactive operations given to you must be perfectly reproduced in the generated HTML, meaning they must have complete functionality, and after completion the page must be consistent with the corresponding screenshot.
\end{enumerate}

\vspace{1mm}

Please use the following libraries:

\begin{itemize}
    \item \texttt{React 18}: \url{https://cdn.jsdelivr.net/npm/react@18.0.0/umd/react.development.js}
    \item \texttt{ReactDOM 18}: \url{https://cdn.jsdelivr.net/npm/react-dom@18.0.0/umd/react-dom.development.js}
    \item \texttt{Babel}: \url{https://cdn.jsdelivr.net/npm/@babel/standalone/babel.js}
    \item \texttt{Tailwind CSS}: \url{https://cdn.tailwindcss.com}
    \item \texttt{Font Awesome}: \url{https://cdnjs.cloudflare.com/ajax/libs/font-awesome/5.15.3/css/all.min.css}
\end{itemize}

\vspace{1mm}

\textbf{Code Output Format}
\begin{enumerate}
    \item Only output the code inside the complete \texttt{<html></html>} tags.
    \item Ensure the code is complete HTML webpage code that can be rendered directly, and do not omit anything with ellipses.
    \item Do not add markdown, \texttt{html}, or any additional text before or after the code.
\end{enumerate}

\vspace{1mm}

\textbf{Video frames information}

\textbf{<VIDEO FRAMES INFORMATION>}.

\end{tcolorbox}
\caption{Prompt used to reconstruct webpage HTML code from clipped webpage interaction video frames}
\label{fig:generation-template}
\end{figure*}

\begin{figure}[t]
\centering
\begin{tcolorbox}[
    colback=gray!2,
    colframe=blue!70!black,
    colbacktitle=blue!12!white,
    title=\textbf{Video-based Webpage Reconstruction Prompt for Baseline Inference},
    fonttitle=\bfseries,
    coltitle=blue!50!black,
    enhanced,
    sharp corners,
    boxrule=0.5pt,
    left=5pt,
    right=5pt,
    top=5pt,
    bottom=5pt,
    titlerule=0mm,
    width=0.5\textwidth,
    title style={left color=blue!8!white, right color=blue!5!white},
]
\small
You are highly skilled in building interactive web pages using React and Tailwind, and you can precisely reconstruct complete, interactive HTML web pages based on videos provided by the user.

\vspace{1mm}

\textbf{UI Requirements}
\begin{enumerate}
    \item Do not omit any details. Background colors, fonts, font sizes, spacing, borders, icons, text, etc., must strictly match the video.
    \item Every single line of text in the video must be presented exactly as it is.
    \item For image content, please use real images from the \url{https://picsum.photos/} library, with URLs similar to \url{https://picsum.photos/id/.../.../...}. Each image must explicitly list its URL; do not use reusable image components. The image URL for each web component must be fixed and not randomly regenerated every time.
\end{enumerate}

\vspace{1mm}

\textbf{Core Task Requirements}

Please apply a ``microscopic'' level of observation to achieve a 1:1 pixel-perfect reproduction of the web page interactions:

\begin{enumerate}
    \item \textbf{Observe overall page changes}: Carefully watch the interactive web page video sent by the user, and do not miss any details on the page, such as pop-up notifications appearing in the top right corner, newly added list items, etc.
    \item All interactive operations shown to you must be perfectly replicated in the generated HTML. This means it must be fully functional, and upon completion, the page must exactly match the content in the video.
\end{enumerate}

\vspace{1mm}

\textbf{Code Output Format}
\begin{enumerate}
    \item Only output the complete code within the \texttt{<html></html>} tags.
    \item Ensure the code is a complete, directly renderable HTML webpage. Do not omit anything using ellipses.
    \item Do not add markdown backticks, \texttt{html}, or any additional text before or after the code.
\end{enumerate}

\end{tcolorbox}
\caption{Prompt used to reconstruct interactive webpage HTML code directly from videos.}
\label{fig:prompt_video_baseline}
\end{figure}

\begin{figure}[t]
\centering
\begin{tcolorbox}[
    colback=gray!2,
    colframe=blue!70!black,
    colbacktitle=blue!12!white,
    title=\textbf{Frame-based Webpage Reconstruction Prompt for Baseline Inference},
    fonttitle=\bfseries,
    coltitle=blue!50!black,
    enhanced,
    sharp corners,
    boxrule=0.5pt,
    left=5pt,
    right=5pt,
    top=5pt,
    bottom=5pt,
    titlerule=0mm,
    width=0.5\textwidth,
    title style={left color=blue!8!white, right color=blue!5!white},
]
\small
You are highly skilled in building interactive web pages using React and Tailwind, and you can precisely reconstruct complete, interactive HTML web pages based on frames extracted from videos provided by the user.

\vspace{2mm}

\vspace{2mm}

\textbf{UI Requirements}
\begin{enumerate}
    \item Do not omit any details. Background colors, fonts, font sizes, spacing, borders, icons, text, etc., must strictly match the extracted frames.
    \item Every single line of text in the extracted frames must be presented exactly as it is.
    \item For image content, please use real images from the \url{https://picsum.photos/} library, with URLs similar to \url{https://picsum.photos/id/.../.../}. Each image must explicitly list its URL; do not use reusable image components. The image URL for each web component must be fixed and not randomly regenerated every time.
\end{enumerate}

\vspace{2mm}

\textbf{Core Task Requirements}

Please apply a ``microscopic'' level of observation to achieve a 1:1 pixel-perfect reproduction of the web page interactions.

Carefully watch the chronologically ordered frames extracted from the interactive web page video sent by the user, and do not miss any details on the page, such as pop-up notifications appearing in the top right corner, newly added list items, and other subtle visual or content changes.

All interactive operations shown to you must be perfectly replicated in the generated HTML. This means the webpage must be fully functional, and upon completion, the page must exactly match the content in the extracted frames.

\vspace{2mm}

\textbf{Code Output Format}
\begin{enumerate}
    \item Only output the complete code within the \texttt{<html></html>} tags.
    \item Ensure the code is a complete, directly renderable HTML webpage. Do not omit anything using ellipses.
    \item Do not add markdown backticks, \texttt{html}, or any text before or after the code.
\end{enumerate}
\end{tcolorbox}
\caption{Prompt used to reconstruct interactive webpage HTML code directly from videos frames.}
\label{fig:prompt_frame_baseline}
\end{figure}

% Add this in the preamble:
% \usepackage{enumitem}

\begin{figure*}[t]
\centering
\begin{tcolorbox}[
    colback=gray!2,
    colframe=blue!70!black,
    colbacktitle=blue!12!white,
    title=\textbf{Interaction Selection Prompt for Evaluation},
    fonttitle=\bfseries,
    coltitle=blue!50!black,
    enhanced,
    sharp corners,
    boxrule=0.5pt,
    left=5pt,
    right=5pt,
    top=5pt,
    bottom=5pt,
    titlerule=0mm,
    width=1\textwidth,
    title style={left color=blue!8!white, right color=blue!5!white},
]
\small
You are a webpage interaction replay assistant. I will give you a set of video frame screenshots arranged in chronological order, capturing a user interaction on a webpage.
I will also provide a DOM Tree and current screenshot of a replicated webpage.

\vspace{1mm}

\textbf{DOM Tree Key Fields}

Each interactable element contains: id (identifier used in action instructions), tag, attrs, visible\_text, address (\texttt{"zone--[x,y,w,h]"}), can\_interact, input\_value, options, etc.

\vspace{1mm}

\textbf{Your Task}
\begin{enumerate}[leftmargin=*, itemsep=0pt, topsep=1pt, parsep=0pt, partopsep=0pt]
    \item Compare the frames to identify the interaction that occurred in the original webpage.
    \item The interaction is usually exactly one of: \textbf{click}, \textbf{enter} (typing), \textbf{select} (dropdown), or \textbf{scroll}.
    \item Locate the corresponding element in the DOM Tree by matching visible\_text, address, tag, input\_value, options, and surrounding context.
    \item Output the action instruction needed to replay the interaction on the replicated webpage.
\end{enumerate}

\vspace{1mm}

\textbf{Important Notes on Number of Actions}
\begin{itemize}[leftmargin=*, itemsep=0pt, topsep=1pt, parsep=0pt, partopsep=0pt]
    \item In almost all cases, the interaction can and should be reproduced with \textbf{one single action}.
    \item Only output \textbf{two actions} in the special case where the video shows a dropdown/select-like interaction, but the replicated DOM Tree does not contain a usable \texttt{select} element or matching selectable option.
    \item In that special case, you may reproduce the interaction with two \texttt{click} actions, for example: first click to open the dropdown, then click the target option.
    \item Do \textbf{not} output multiple actions for ordinary clicks, typing, scrolling, or standard select elements.
    \item Prefer a single \texttt{select[element\_id][option\_text]} whenever a proper select/dropdown option exists in the DOM Tree.
\end{itemize}

\vspace{1mm}

\textbf{Important Notes on Missing Operations}
\begin{itemize}[leftmargin=*, itemsep=0pt, topsep=1pt, parsep=0pt, partopsep=0pt]
    \item If the interaction shown in the video cannot be matched to any actionable element in either the current screenshot or the DOM Tree, output \texttt{\textbackslash boxed\{\{None\}\}}.
    \item Use \texttt{None} only when no reasonable click, enter, select, or scroll action can reproduce the observed interaction on the replicated webpage.
\end{itemize}

\vspace{1mm}

\textbf{Important Notes on Scroll}
\begin{itemize}[leftmargin=*, itemsep=0pt, topsep=1pt, parsep=0pt, partopsep=0pt]
    \item If the interaction is a scroll, it is \textbf{very likely a scroll to the bottom of the page}---use \texttt{scroll[]} with no element id.
    \item Only use \texttt{scroll[element\_id]} if the frames clearly show scrolling to a specific element rather than to the page bottom.
\end{itemize}

\vspace{1mm}

\textbf{Output Format}

Wrap the final action instruction in LaTeX \texttt{\textbackslash boxed\{\{\}\}}.

Allowed formats:
\begin{itemize}[leftmargin=*, itemsep=0pt, topsep=1pt, parsep=0pt, partopsep=0pt]
    \item \texttt{click[element\_id]}
    \item \texttt{enter[element\_id][input\_content]}
    \item \texttt{select[element\_id][option\_text]}
    \item \texttt{scroll[element\_id]} --- scroll to bring a specific element into view
    \item \texttt{scroll[]} --- scroll all the way to the bottom of the page
    \item \texttt{click[element\_id\_1]; click[element\_id\_2]} --- only for the special dropdown/select replacement case described above
    \item \texttt{None} --- only when the interaction cannot be matched to the screenshot or DOM Tree
\end{itemize}

Briefly explain your reasoning first, then give the final answer.
The final answer must contain exactly one \texttt{\textbackslash boxed\{\{\}\}}.
If two clicks are needed, put both click actions inside the same \texttt{\textbackslash boxed\{\{\}\}}.
If no valid operation can be found, output exactly \texttt{\textbackslash boxed\{\{None\}\}}.

\vspace{1mm}

\textbf{DOM Tree}

\textbf{<DOM\_TREE>}

\vspace{1mm}

\textbf{Input Visual Observations}

The following are the video frame sequence and replicated webpage screenshot:

\textbf{<VIDEO\_FRAME\_SEQUENCE\_AND\_REPLICATED\_WEBPAGE\_SCREENSHOT>}

\end{tcolorbox}
\caption{Interaction replay prompt used to infer executable action instructions from video frames, DOM trees, and replicated webpage screenshots.}
\label{fig:action_replay_prompt}
\end{figure*}

\begin{figure*}[t]
\centering
\begin{tcolorbox}[
    colback=gray!2,
    colframe=blue!70!black,
    colbacktitle=blue!12!white,
    title=\textbf{Interaction Verification Prompt for Evaluation},
    fonttitle=\bfseries,
    coltitle=blue!50!black,
    enhanced,
    sharp corners,
    boxrule=0.5pt,
    left=4pt,
    right=4pt,
    top=3pt,
    bottom=3pt,
    titlerule=0mm,
    width=0.98\textwidth,
    title style={left color=blue!8!white, right color=blue!5!white},
    before upper={
        \setlength{\parskip}{0pt}
        \setlength{\itemsep}{0.5pt}
        \setlength{\parsep}{0pt}
        \setlength{\topsep}{1pt}
        \setlength{\partopsep}{0pt}
        \linespread{0.96}\selectfont
    }
]
\footnotesize

You are evaluating whether a web interaction was correctly replicated by comparing \textbf{page visual similarity}.

The two groups of frames below each capture a complete interaction process including before, during, and after the action.

\begin{itemize}[leftmargin=*, itemsep=0.5pt, topsep=1pt, parsep=0pt, partopsep=0pt]
    \item Group 1 (Expected): Original webpage recording
    \item Group 2 (Actual): Replicated webpage after attempting the same interaction
\end{itemize}

The operation being performed is: \textbf{<ACTION INFORMATION>}

\vspace{1mm}

\textbf{Step 1: Identify the Operation Type}

Determine whether the operation is:
\begin{itemize}[leftmargin=*, itemsep=0.5pt, topsep=1pt, parsep=0pt, partopsep=0pt]
    \item \textbf{Click / Enter / Select}: An action that changes a specific UI element's state
    \item \textbf{Scroll}: An action whose purpose is to reveal page content
\end{itemize}

\vspace{1mm}

\textbf{Step 2: Visual Score}

Output an integer from 0--100, or -1 if incomparable.

\vspace{1mm}

\textbf{For Click, Enter, or Select operations:}
\begin{itemize}[leftmargin=*, itemsep=0.5pt, topsep=1pt, parsep=0pt, partopsep=0pt]
    \item Compare the \textbf{final post-action page state} between Group 1 and Group 2.
    \item Ignore any differences in the initial page state before the action; only the result matters.
    \item Score how visually similar the two pages look \textbf{after} the operation completes.
    \item Output -1 if the two post-action pages are so fundamentally different in layout or content that no meaningful visual comparison is possible, such as completely different pages being loaded or drastically different structures.
\end{itemize}

\vspace{1mm}

\textbf{For Scroll operations:}
\begin{itemize}[leftmargin=*, itemsep=0.5pt, topsep=1pt, parsep=0pt, partopsep=0pt]
    \item Scrolling's sole purpose is to make content visible. Do not evaluate whether a scroll motion occurred.
    \item Compare the \textbf{total visible content} across all frames in each group.
    \item If Group 1 required scrolling to reveal content, but Group 2 already shows all that content without scrolling, this is a \textbf{successful match}. Score based on content similarity between what is visible in both groups.
    \item Focus on whether the same page content is accessible or visible, not on the scrolling gesture itself.
\end{itemize}

\vspace{1mm}

\textbf{Special Case --- Pre-executed State:}

If the action was already in the ``done'' state before it was attempted, such as a checkbox already checked, a tab already active, or a value already set, the operation had no visible effect in Group 2. In this case:
\begin{itemize}[leftmargin=*, itemsep=0.5pt, topsep=1pt, parsep=0pt, partopsep=0pt]
    \item The functional score will be ``failed'' because no change occurred.
    \item The \textbf{visual score should still reflect the similarity of the page states normally}. The pages can still look identical even without an observable change.
\end{itemize}

\vspace{1mm}

\textbf{Scoring criteria:}
\begin{itemize}[leftmargin=*, itemsep=0.5pt, topsep=1pt, parsep=0pt, partopsep=0pt]
    \item 90--100: Post-action pages look visually identical or nearly identical.
    \item 70--89: Post-action pages are very similar with only minor visual differences.
    \item 50--69: Post-action pages are similar but with noticeable differences.
    \item 20--49: Post-action pages differ significantly.
    \item 0--19: Post-action pages look completely different or an error state appeared.
    \item -1: The two post-action pages are fundamentally incomparable, such as different pages or broken layouts.
\end{itemize}

\vspace{1mm}

\textbf{Step 3: Functional Score}

Determine whether the operation was actually executed in Group 2, and briefly explain the reason:
\begin{itemize}[leftmargin=*, itemsep=0.5pt, topsep=1pt, parsep=0pt, partopsep=0pt]
    \item \texttt{success}: The operation clearly took effect, or the target state was already achieved before the action.
    \item \texttt{failed}: The operation did not take effect or produced an unexpected error state.
\end{itemize}

When making this judgment, explicitly mention the key visual evidence, such as:
\begin{itemize}[leftmargin=*, itemsep=0.5pt, topsep=1pt, parsep=0pt, partopsep=0pt]
    \item whether the expected UI state change appeared;
    \item whether the target element changed as intended;
    \item whether the page navigated, updated, expanded, selected, or scrolled as expected;
    \item whether the action produced no visible effect or an error state.
\end{itemize}

\vspace{1mm}

\textbf{Output Format}

Briefly explain your reasoning first, \textbf{ensuring that the justification of the functional score is included}, then provide both scores on separate lines:

\begin{itemize}[leftmargin=*, itemsep=0.5pt, topsep=1pt, parsep=0pt, partopsep=0pt]
    \item Visual score: \texttt{\textbackslash score\{N\}}, where N is an integer from 0 to 100, or \texttt{\textbackslash score\{-1\}} if the pages are incomparable.
    \item Functional score: \texttt{\textbackslash passed\{success\}} or \texttt{\textbackslash passed\{failed\}}.
\end{itemize}
\end{tcolorbox}
\caption{Prompt used to evaluate whether a replicated webpage correctly preserves the visual and functional effects of an interaction.}
\label{fig:visual-similarity-template}
\end{figure*}

\begin{figure*}[t]
\centering
\begin{tcolorbox}[
    colback=gray!2,
    colframe=blue!70!black,
    colbacktitle=blue!12!white,
    title=\textbf{Initial State Visual Similarity Prompt for Evaluation},
    fonttitle=\bfseries,
    coltitle=blue!50!black,
    enhanced,
    sharp corners,
    boxrule=0.5pt,
    left=4pt,
    right=4pt,
    top=3pt,
    bottom=3pt,
    titlerule=0mm,
    width=0.98\textwidth,
    title style={left color=blue!8!white, right color=blue!5!white},
    before upper={
        \setlength{\parskip}{0pt}
        \setlength{\itemsep}{0.5pt}
        \setlength{\parsep}{0pt}
        \setlength{\topsep}{1pt}
        \setlength{\partopsep}{0pt}
        \linespread{0.96}\selectfont
    }
]
\footnotesize

You are evaluating the visual similarity between two webpage screenshots.

\begin{itemize}[leftmargin=*, itemsep=0.5pt, topsep=1pt, parsep=0pt, partopsep=0pt]
    \item Image 1: The first frame from the original webpage recording (reference)
    \item Image 2: A screenshot of the replicated/rendered webpage's initial state
\end{itemize}

\vspace{1mm}

\textbf{Your Task}

Compare these two webpage screenshots and score their visual similarity.

\vspace{1mm}

\textbf{Focus on:}
\begin{itemize}[leftmargin=*, itemsep=0.5pt, topsep=1pt, parsep=0pt, partopsep=0pt]
    \item Overall layout and structure
    \item Color scheme and visual style
    \item Content placement and hierarchy
    \item Key UI elements, including headers, navigation, main content area, and hero sections
\end{itemize}

\vspace{1mm}

\textbf{Scoring criteria:}
\begin{itemize}[leftmargin=*, itemsep=0.5pt, topsep=1pt, parsep=0pt, partopsep=0pt]
    \item 90--100: The pages look visually identical or nearly identical.
    \item 70--89: Very similar with only minor visual differences, such as slight color variation or minor spacing differences.
    \item 50--69: Similar but with noticeable differences.
    \item 20--49: Differ significantly in layout or content.
    \item 0--19: Look completely different.
\end{itemize}

\vspace{1mm}

\textbf{Output Format}

Briefly explain your reasoning, then provide the score on a separate line:

\begin{itemize}[leftmargin=*, itemsep=0.5pt, topsep=1pt, parsep=0pt, partopsep=0pt]
    \item Visual similarity score: \texttt{\textbackslash score\{N\}}, where N is an integer from 0 to 100.
\end{itemize}

\end{tcolorbox}
\caption{Prompt used to evaluate the initial-state visual similarity between the original webpage recording and the replicated webpage.}
\label{fig:initial-state-visual-similarity-template}
\end{figure*}

\subsection{Error Analysis}

We provide a qualitative error analysis to better understand the remaining challenges in UI video-to-code generation. We manually inspect 220 failed cases from \textit{WebVideo2Code-Real} and categorize them into five error types, as summarized in Table~\ref{tab:error_analysis}. The dominant failure mode is \textit{event binding and interaction logic errors}, which account for 64.1\% of the inspected failures. This suggests that the main bottleneck of UI video-to-code generation is not only reconstructing webpage appearance, but also correctly implementing executable behavior from action-conditioned state transitions.

\begin{table}[t]
\centering
% \small
\renewcommand{\arraystretch}{1.1}
\setlength{\tabcolsep}{3.5pt}
\caption{
Error distribution on manually inspected failed cases from \textit{WebVideo2Code-Real}.
}
\begin{tabular}{lc}
\toprule
Error Type & \ \ \ \ \ \ \ \ Ratio \\
\midrule
Event binding \& logic failures & \ \ \ \ \ \ \ \  64.1\% \\
Structural \& systemic errors & \ \ \ \ \ \ \ \  24.1\% \\
HTML render errors  & \ \ \ \ \ \ \ \  5.9\% \\
Visual \& layout anomalies & \ \ \ \ \ \ \ \ 5.0\% \\
Unknown & \ \ \ \ \ \ \ \  0.9\% \\
\bottomrule
\end{tabular}
\label{tab:error_analysis}
\vspace{-4mm}
\end{table}

\vpara{Event binding and logic failures.} The most common failure occurs when the generated webpage contains visually plausible interactive components but fails to implement the correct behavior. For example, a button, tab, dropdown, or selection widget may be present, but clicking or selecting it does not trigger the expected state change, or updates the wrong region of the page. These errors indicate that behavior-faithful UI video-to-code generation requires more than identifying interactive elements: the model must also bind each demonstrated action to the correct event handler, state update, and post-action UI state.

\vpara{Structural and systemic errors.} Some failures arise from incorrect or incomplete webpage structure. In these cases, the generated implementation may miss important components, organize sections in the wrong hierarchy, or fail to preserve the global relationship among UI states. Such structural errors often affect multiple interactions at once, because later actions depend on page elements or states that were not correctly reconstructed.

\vpara{HTML render errors.} A smaller portion of failures are caused by invalid or unstable generated code. The generated HTML/CSS/JavaScript may fail to render, produce a blank page, or break during browser execution. These cases directly prevent browser-based interaction verification and highlight the need for stronger executable-code constraints during generation.

\vpara{Visual and layout anomalies.} Some generated webpages are executable but visually inconsistent with the target video. Typical issues include incorrect spacing, misaligned components, missing visual details, inconsistent typography, or wrong component placement. Although these errors may not always prevent interaction execution, they reduce visual similarity and can make the generated post-action state difficult to match with the target transition.

\vpara{Unknown failures.} A small portion of failures, 0.9\%, are categorized as unknown. These cases are difficult to attribute to a single cause, either because multiple error sources occur simultaneously or because the recorded execution trace does not provide enough evidence to identify the primary failure mode.

\subsection{Demo Cases}

\begin{figure*}
    \centering
    \includegraphics[width=0.9\linewidth]{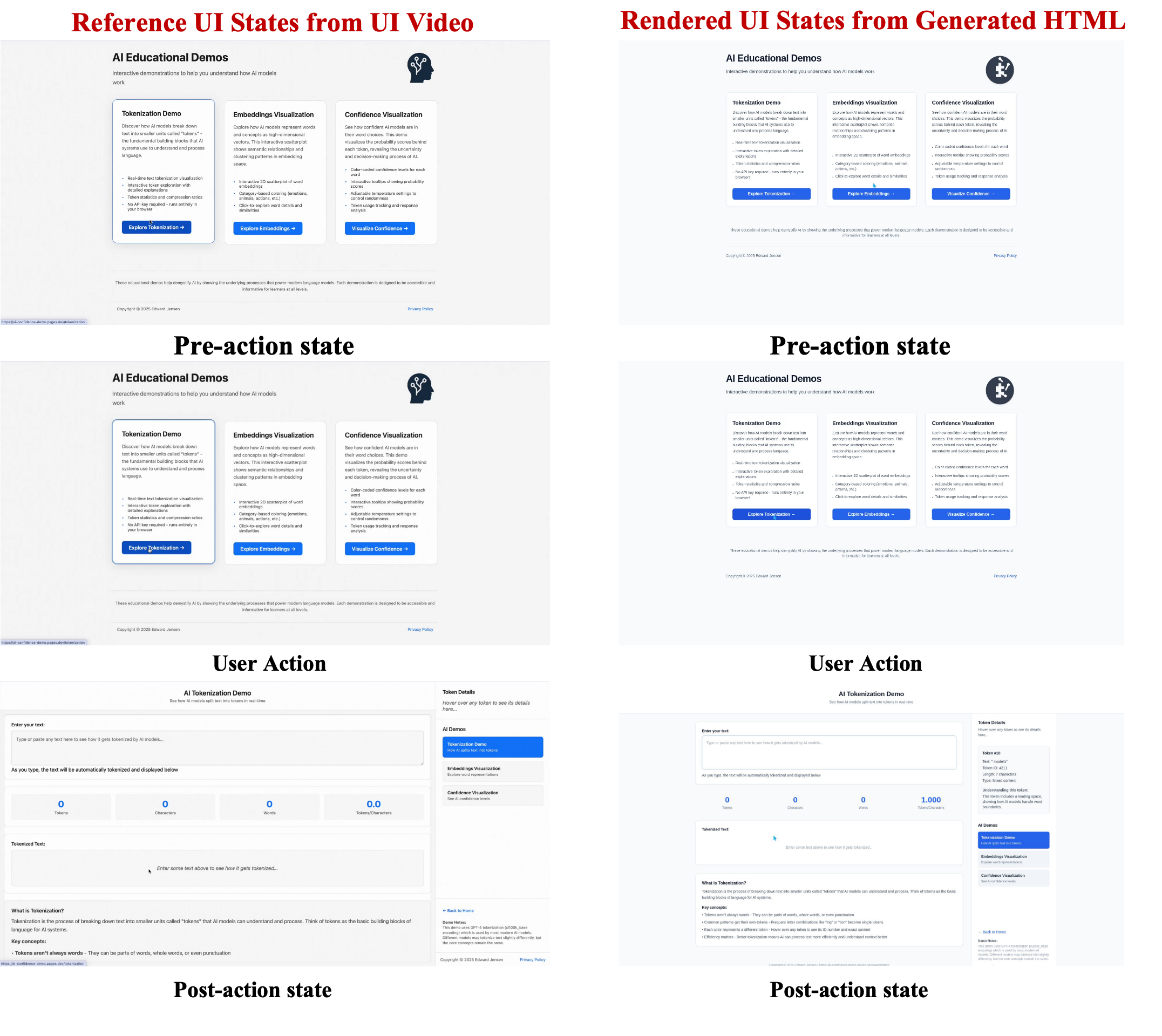}
    \caption{Demo cases of Video2Code on click interactions.}
    \label{fig:demo_click}
\end{figure*}

\begin{figure*}
    \centering
    \includegraphics[width=0.9\linewidth]{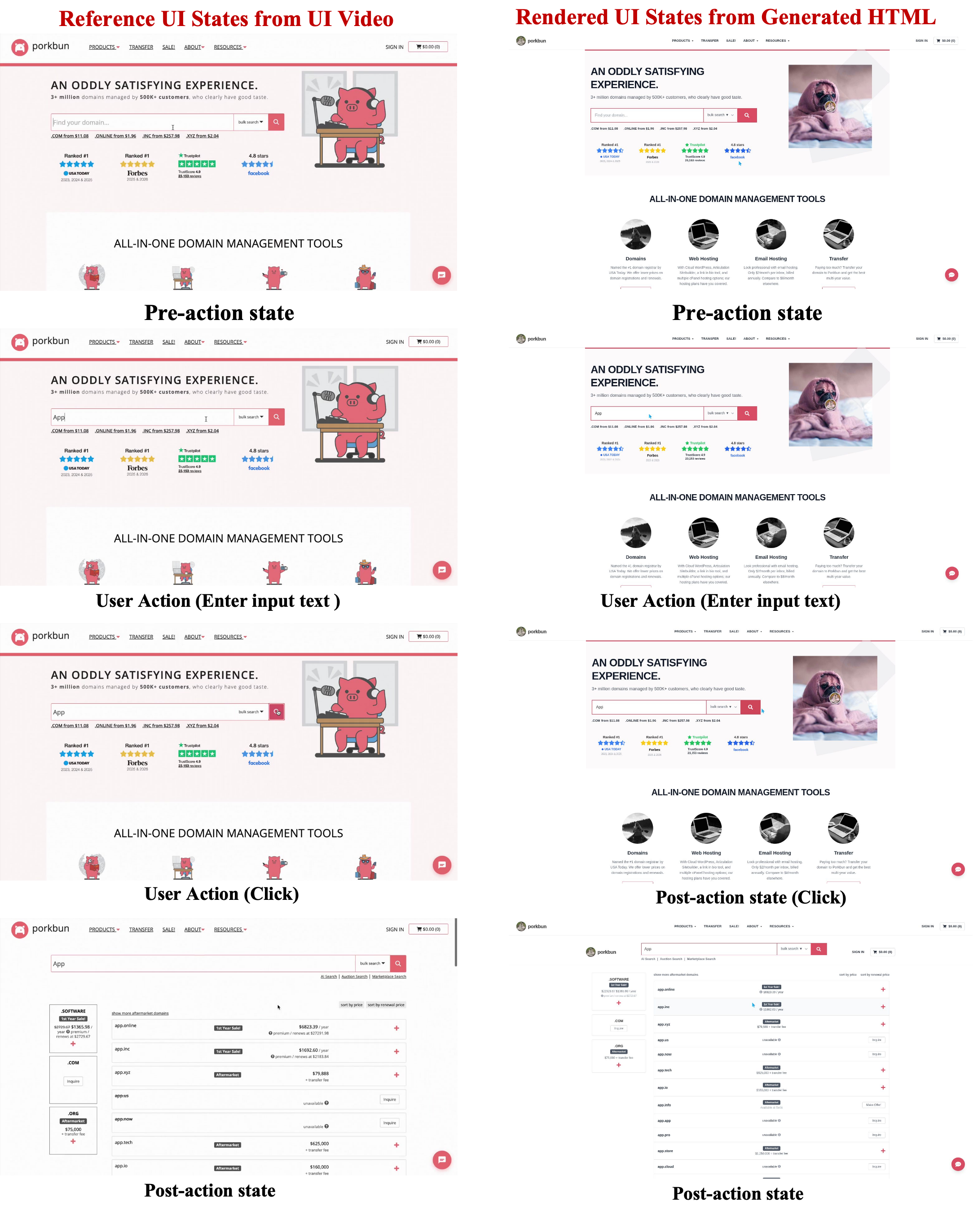}
    \caption{Demo cases of Video2Code on a text-input and click interaction sequence.}
    \label{fig:demo_text_input}
\end{figure*}

\begin{figure*}
    \centering
    \includegraphics[width=0.8\linewidth]{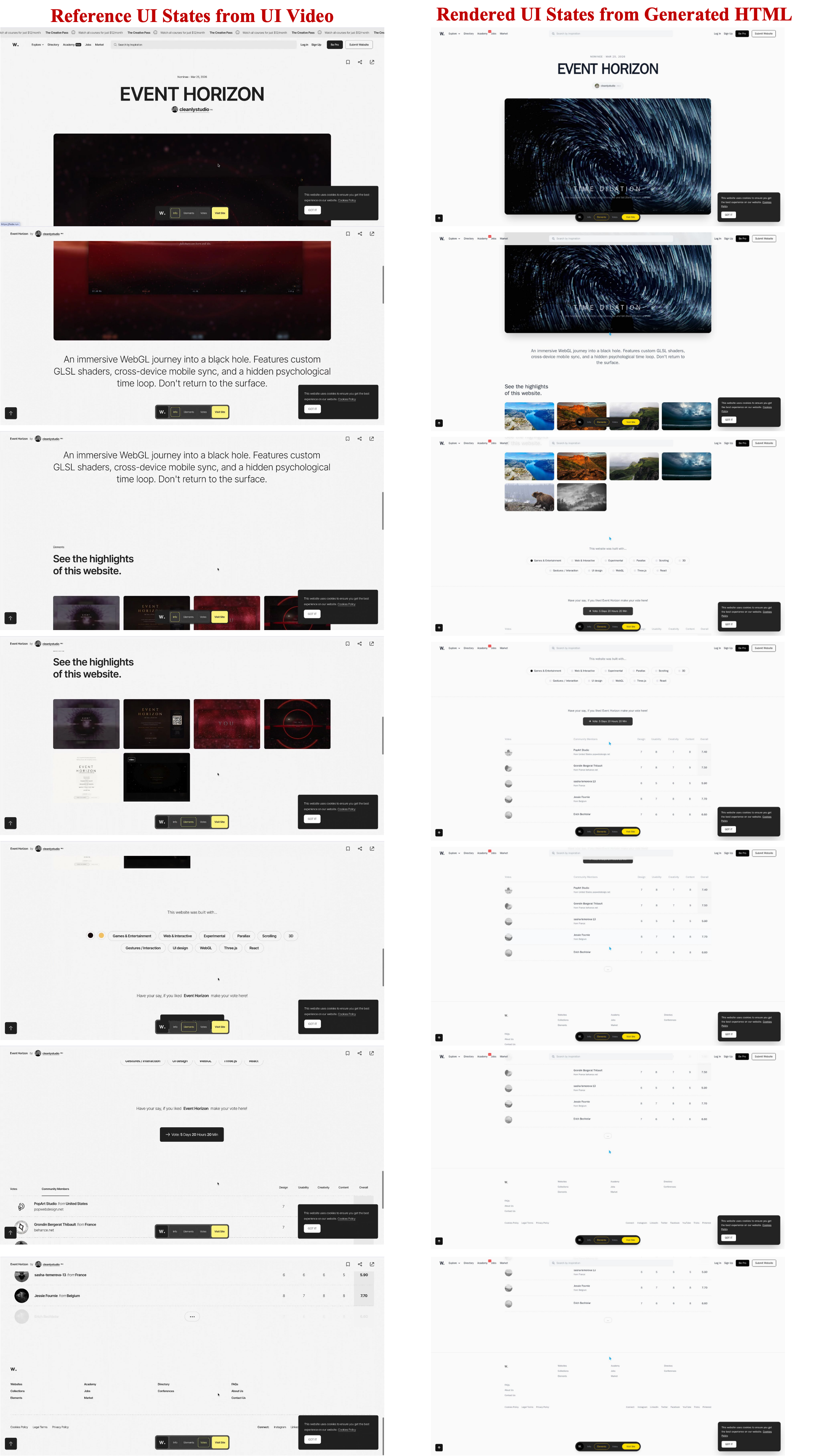}
    \caption{Demo cases of Video2Code on scroll interactions. }
    \label{fig:demo_scroll}
\end{figure*}

\begin{figure*}
    \centering
    \includegraphics[width=0.8\linewidth]{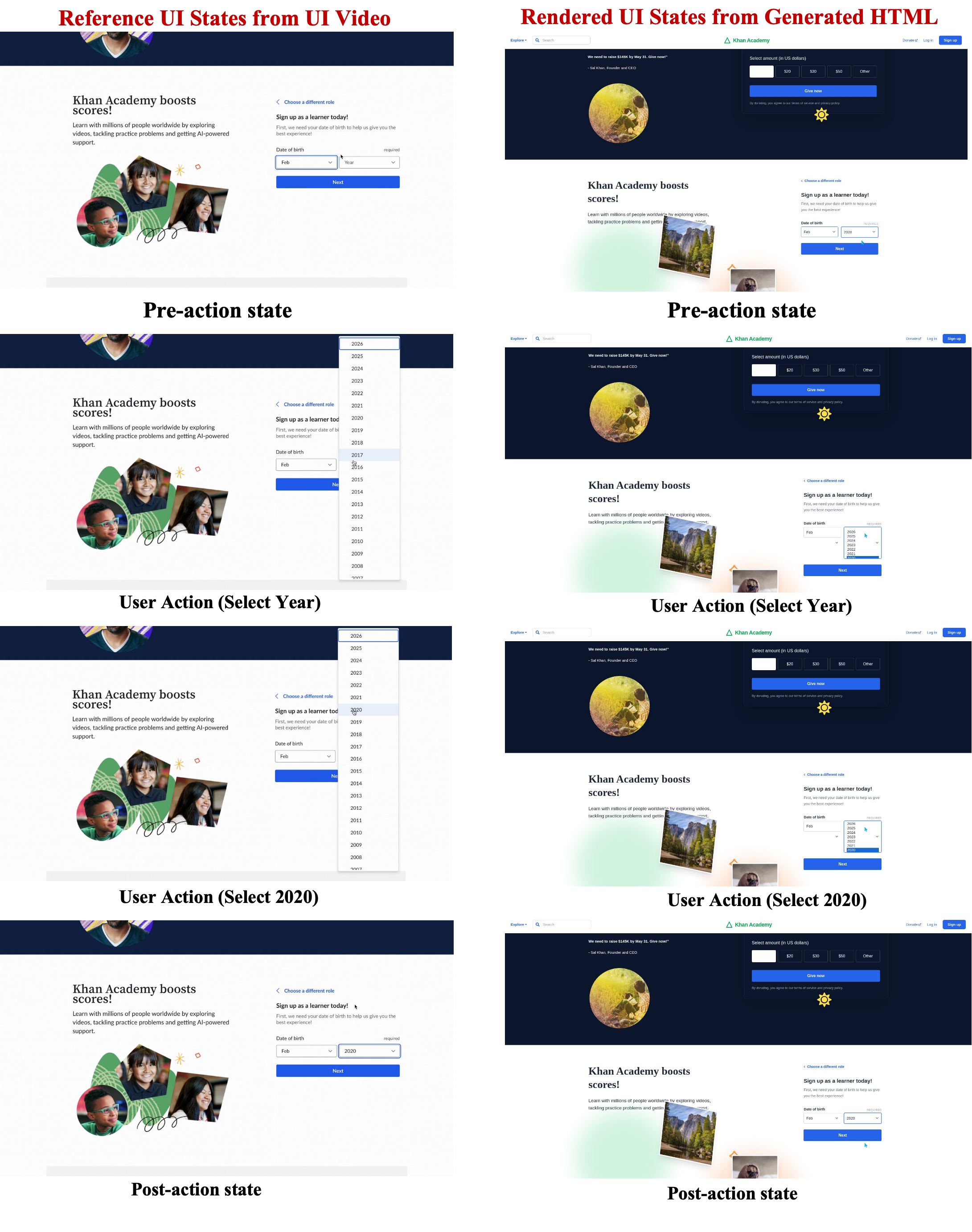}
    \caption{Demo cases of Video2Code on selection interactions.}
    \label{fig:demo_select}
\end{figure*}

We present qualitative examples of UI video-to-code generation with Video2Code. For each example, we visualize the action-critical segment revisited by Video2Code and the rendered states of the generated webpage before and after executing the corresponding user action. These examples cover representative interaction patterns, including click, text input, scroll, and selection, and illustrate how Video2Code uses focused temporal revisit to recover local state-action-state evidence and translate it into executable webpage behavior. Figures~\ref{fig:demo_click}--\ref{fig:demo_select} show representative cases for these action types, demonstrating behaviors such as click-triggered updates, input-dependent changes, scroll-dependent state transitions, and selection-based content updates.

\begin{figure*}
    \centering
    \includegraphics[width=0.6\linewidth]{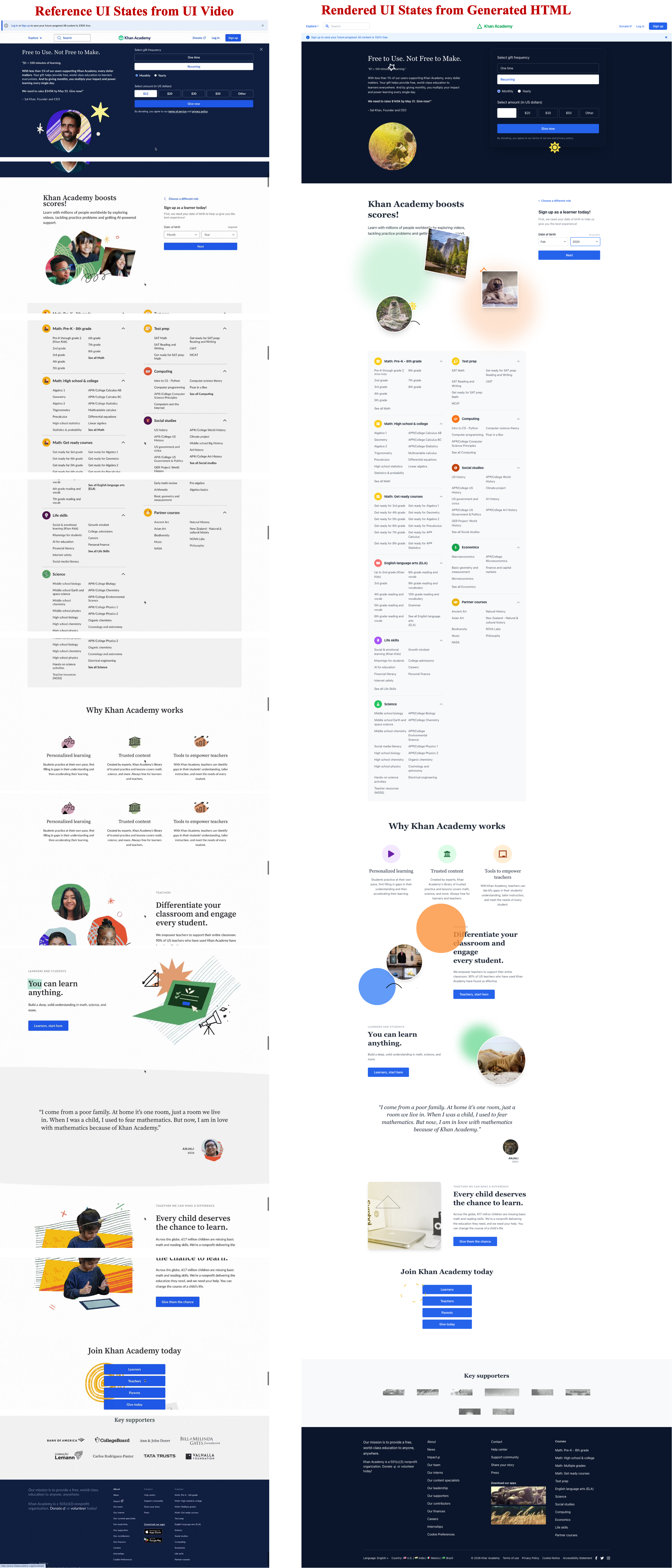}
    \caption{Demo case of Video2Code on a scroll interaction. 
    The figure compares the target frames from the input video with the corresponding full-page rendered states of the generated webpage, showing that Video2Code recovers the scroll-dependent content transition.}
    \label{fig:demo_5}
\end{figure*}

\begin{figure*}
    \centering
    \includegraphics[width=0.85\linewidth]{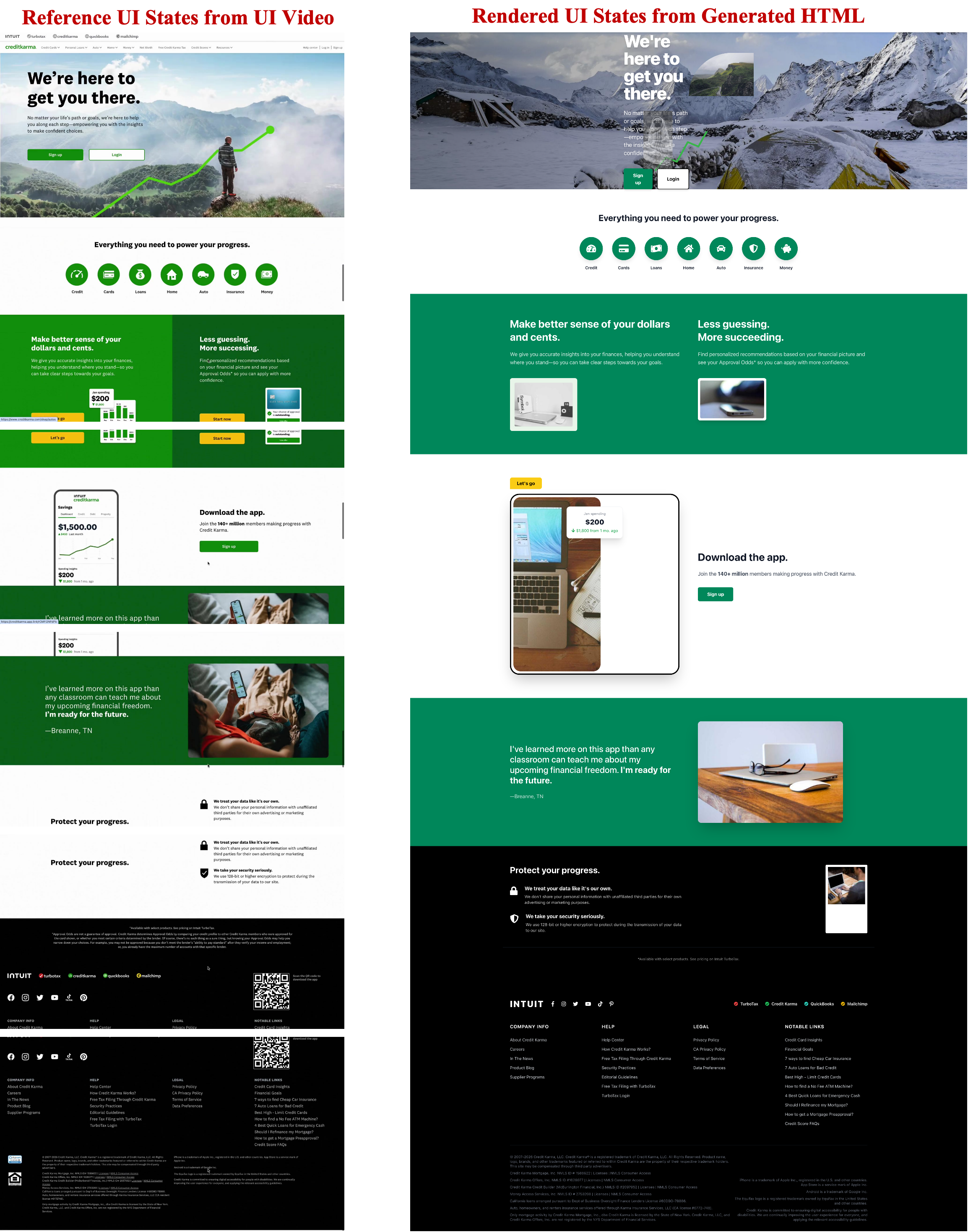}
    \caption{Demo case of Video2Code on a scroll interaction. 
    The generated webpage preserves the long-page structure and reaches the target viewport state after executing the demonstrated scroll operation.}
    \label{fig:demo_6}
\end{figure*}

\begin{figure*}
    \centering
    \includegraphics[width=0.85\linewidth]{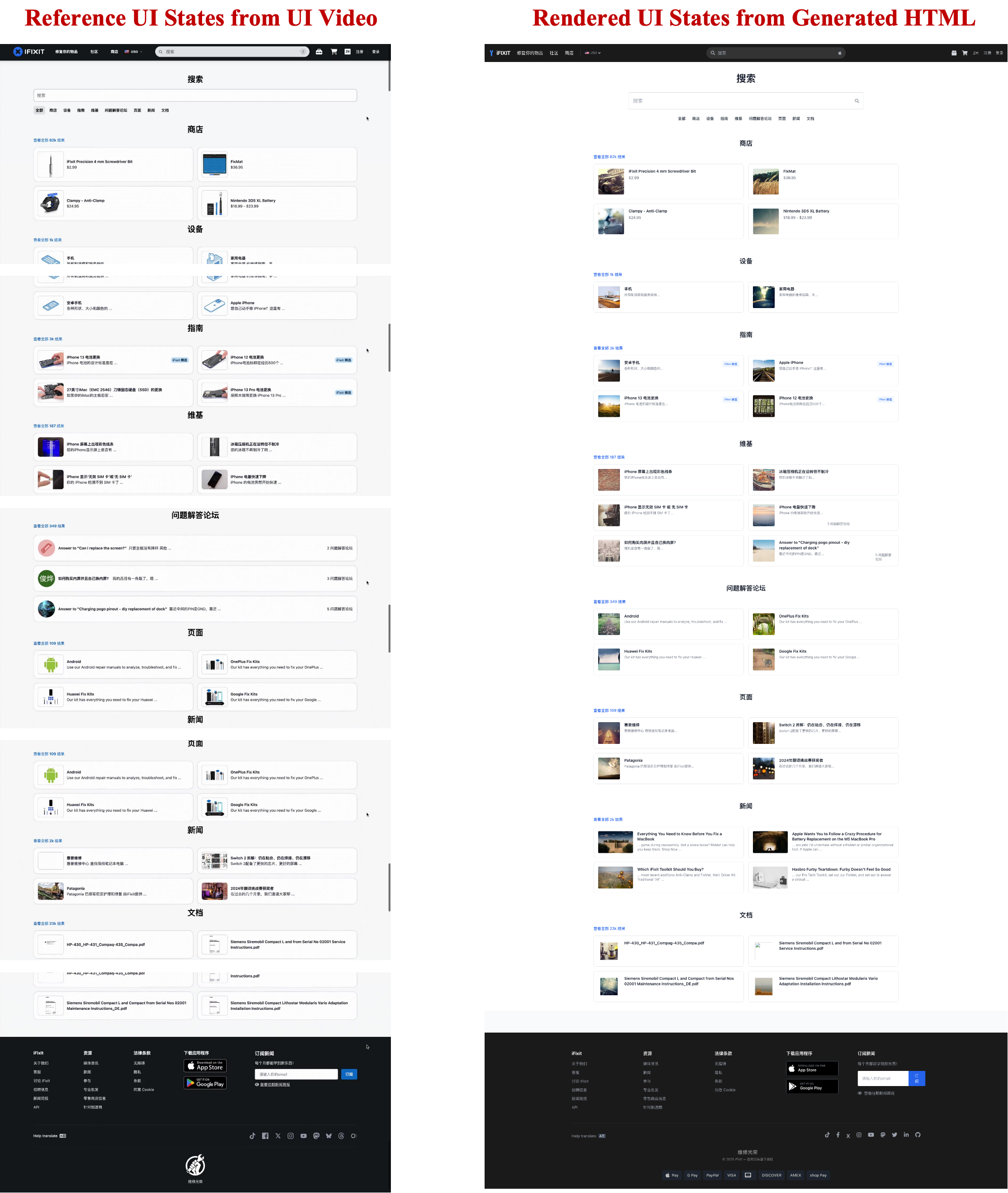}
    \caption{Demo case of Video2Code on a scroll interaction. 
    The target video frames and generated full-screen renderings illustrate how Video2Code reproduces the post-scroll visual state.}
    \label{fig:demo_7}
\end{figure*}

Figures~\ref{fig:demo_5}--\ref{fig:demo_7} show scroll-based demo cases. 
Unlike click or selection interactions, scroll cases are evaluated by comparing the target video frames with the corresponding full-screen rendered states after scrolling. These examples show that Video2Code can preserve long-page layout structure and recover scroll-dependent viewport transitions.

\label{sec:appendix}

\end{document}